\documentclass[lettersize,journal]{IEEEtran}
\usepackage{amsmath,amsfonts}
\usepackage{algorithmic}
\usepackage{algorithm}
\usepackage{array}
\usepackage[caption=false,font=normalsize,labelfont=sf,textfont=sf]{subfig}
\usepackage{textcomp}
\usepackage{stfloats}
\usepackage{url}
\usepackage{verbatim}
\usepackage{graphicx}
\usepackage{cite}

\usepackage{booktabs}
\usepackage{dsfont}
\usepackage{makecell}
\usepackage{multirow}
\usepackage{bbding}
\usepackage{arydshln} 
\usepackage{amsfonts}            
\usepackage{mathrsfs}            

\usepackage{tikz}
\usepackage{comment}
\usepackage{amsmath,amssymb} 
\usepackage{color}
\usepackage{soul}
\usepackage{xcolor}
\makeatletter
\def\ps@IEEEtitlepagestyle{%
  \def\@oddfoot{\mycopyrightnotice}%
  \def\@oddhead{\hbox{}\@IEEEheaderstyle\leftmark\hfil\thepage}\relax
  \def\@evenhead{\@IEEEheaderstyle\thepage\hfil\leftmark\hbox{}}\relax
  \def\@evenfoot{}%
}
\def\mycopyrightnotice{%
  \begin{minipage}{\textwidth}
  \centering \scriptsize
  Copyright~\copyright~2022 IEEE. Personal use of this material is permitted. However, permission to use this material for any other purposes must be obtained from the IEEE by sending an email to pubs-permissions@ieee.org.
  \end{minipage}
}
\makeatother

\begin{document}

\title{Slow Motion Matters: A Slow Motion Enhanced Network for Weakly Supervised Temporal Action Localization}

\author{
  Weiqi Sun, Rui Su, Qian Yu*~\IEEEmembership{Member,~IEEE} and Dong Xu,~\IEEEmembership{Fellow,~IEEE}
\IEEEcompsocitemizethanks{\IEEEcompsocthanksitem  	 
Weiqi Sun and Qian Yu are with Beihang University.
Rui Su is with Shanghai Artificial Intelligence Laboratory.
Dong Xu is with the Department of Computer Science, The University of Hong Kong. The corresponding author is Qian Yu (e-mail:qianyu@buaa.edu.cn). 
}
}

\markboth{Journal of \LaTeX\ Class Files,~Vol.~14, No.~8, August~2021}%
{Shell \MakeLowercase{\textit{et al.}}: A Sample Article Using IEEEtran.cls for IEEE Journals}



\maketitle

\begin{abstract}
  Weakly supervised temporal action localization (WTAL) aims to localize actions in untrimmed videos with only weak supervision information (\textit{e.g.,} video-level labels). Most existing models handle all input videos with a fixed temporal scale. However, such models are not sensitive to actions whose pace of the movements is different from the ``normal" speed, especially slow-motion action instances, which complete the movements with a much slower speed than their counterparts with a ``normal" speed.
  Here arises the slow-motion blurred issue: It is hard to explore salient slow-motion information from videos at normal speed.
  In this paper, we propose a novel framework termed Slow Motion Enhanced Network (SMEN) to improve the ability of a WTAL network by compensating its sensitivity on slow-motion action segments. 
  The proposed SMEN comprises a Mining module and a Localization module. The mining module generates mask to mine slow-motion-related features by utilizing the relationships between the normal motion and slow motion; while the localization module leverages the mined slow-motion features as complementary information to improve the temporal action localization results. 
  Our proposed framework can be easily adapted by existing WTAL networks and enable them be more sensitive to slow-motion actions. 
  Extensive experiments on three benchmarks are conducted, which demonstrate the high performance of our proposed framework.
  
  \end{abstract}
  


\section{Introduction}
\label{sec:intro}

\IEEEPARstart{T}{emporal} action localization (TAL) is an important yet challenging task for video understanding. It aims at localizing the temporal boundaries (\textit{i.e.}, the starting and ending frames) of the actions of interest and recognizing their action categories in untrimmed videos \cite{idrees2017thumos,gaidon2013temporal}. TAL has wide real-world applications such as video surveillance~\cite{9360626} and abnormality alarm~\cite{li2020abnormality} for aged care. Therefore, this task has attracted increasing attention from the research community and embraced great improvements \cite{long2019gaussian,xu2017r,zeng2019graph,chao2018rethinking,shou2016temporal,9416479,6575133,8897018,8852682,9512048} in recent years. However, these fully supervised methods require extensive manual frame-level annotations, which is labor-consuming and time-costing. 

To address this problem, researchers introduced the task of weakly supervised temporal action localization (WTAL). Specifically, instead of using the frame-level temporal annotation, WTAL leverages weaker but cheaper annotations, \textit{e.g.}, video-level labels, during the training process. There are several solutions to the WTAL task. For example, the works in \cite{paul2018w,yuan2019marginalized,lee2020background} formulate this problem as a multiple instance learning (MIL) task~\cite{zhou2004multi} and treat the entire untrimmed video as a bag containing both positive and negative instances, generating class activation sequence (CAS) to obtain the localization results.

\begin{figure*}[!t]
\centering
\subfloat[\small \textrm{Comparison of a normal motion and a slow motion. These two video segments are sampled from the same video at the same sampling rate.}]{\includegraphics[width=0.9\textwidth]{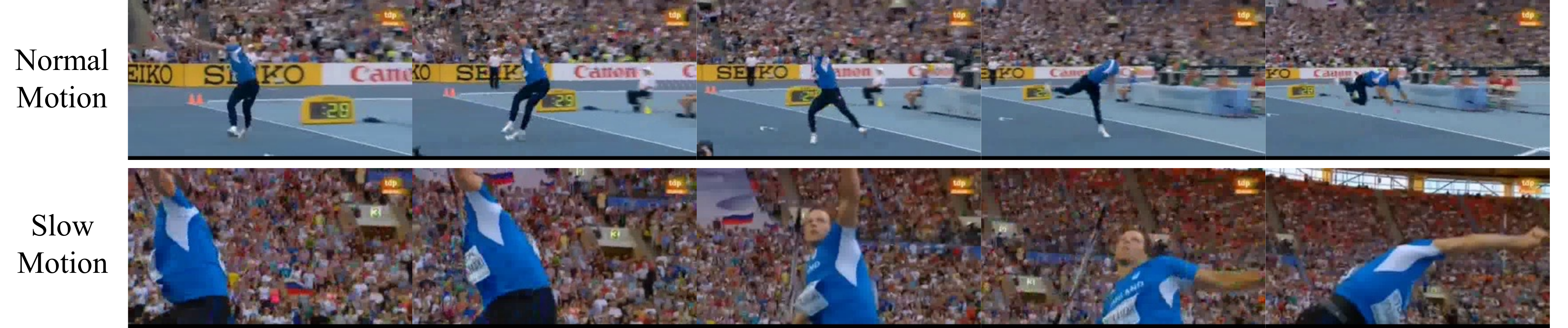}%
\label{fig:m1}}
\hfil
\subfloat[\small \textrm{Comparison of CAS value produced by two models respectively trained on the original video features and sub-sampled features. Baseline network is the same.}]{\includegraphics[width=0.9\textwidth]{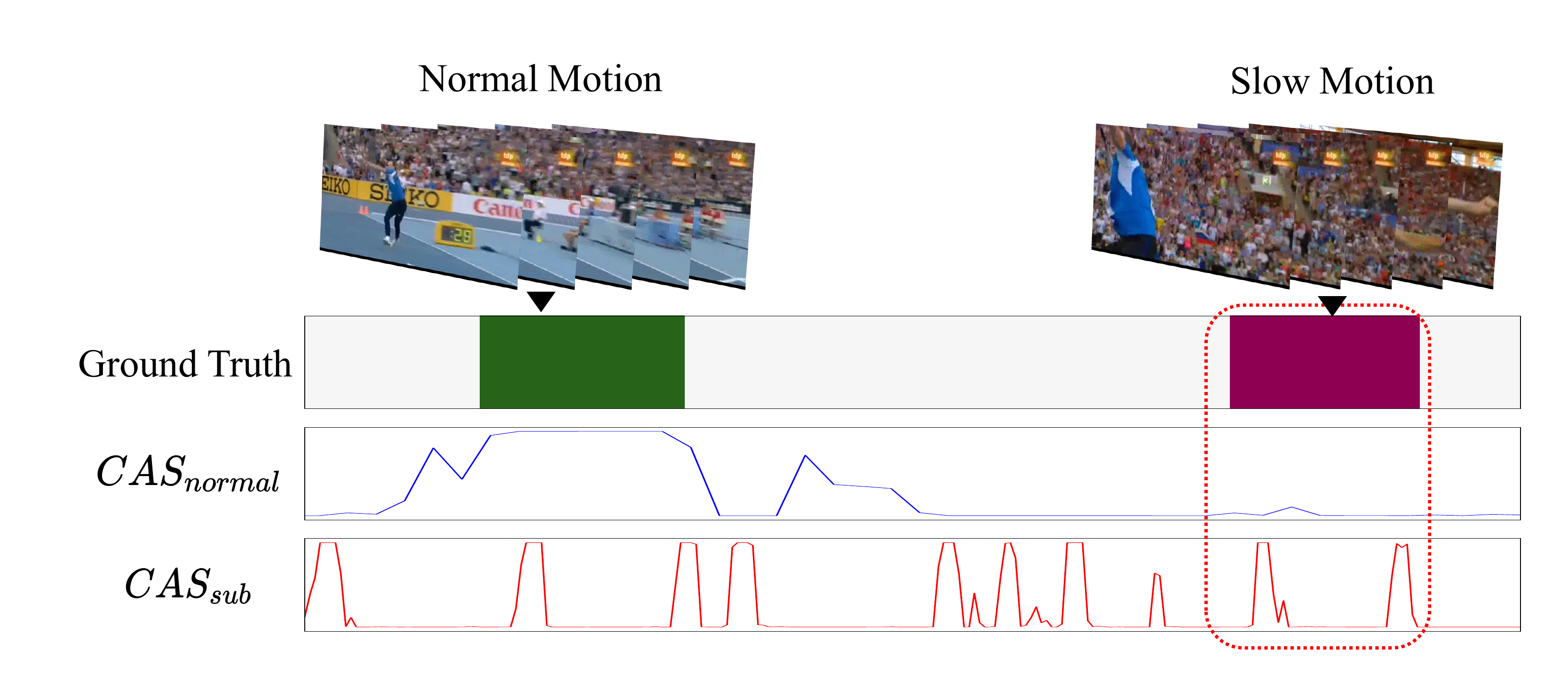}%
\label{fig:m2}}
\caption{An example of action instance of the category ``Javelin Throw'' in the THUMOS'14 dataset. The barcode is the ground truth (GT). Specifically, the green barcode is the action instance with \textit{normal motion}, while the red is the action instance with \textit{slow motion}. The following line charts are CAS value of baseline (ACM-NET~\cite{qu2021acm}) using normal-sampled feature and sub-sampled feature, respectively (\textit{i.e.}, $CAS_{normal}$ and $CAS_{sub}$). The baseline network trained on sub-sampled feature can better handle action instances with slow motion.}
\label{fig:motivation}
\end{figure*}

Although the existing WTAL works~\cite{paul2018w,yuan2019marginalized,lee2020background} have achieved great progress, they overlooked the fact that action instances could exhibit different paces of movement, especially \textit{Slow Motion}, which is defined as an action taking place at a slower than normal speed. Slow motion actions are common, such as the playback in sports videos: In THUMOS'14~\cite{idrees2017thumos} dataset, there are more than 64.0\% videos and 26.4\% action instances containing slow motion. As shown in Fig.~\ref{fig:m1}, the two clips are from the same video. The first clip displays the action ``Javelin Throw" at normal speed, and the second clip shows a slow-motion replay of the action. It is clear to see that a slow-motion action is drastically different from the same action at normal speed. A commonly used pipeline of the existing works is to extract features from video frames sampled at a \textit{fixed} rate and then to process the extracted features to get the final predictions. When determining the sampling rate, researchers mainly consider actions occurring at a normal rate, \textit{i.e.}, normal motion, and ignore the action segments with slow motion. As a result, it is hard for the WTAL frameworks following this pipeline to localize slow motion actions.

Localizing slow motion is not easy. First, unlike normal motion which has salient characteristics, slow motion only has subtle changes in consecutive frames within a temporal period, as shown in Fig.~\ref{fig:m1}. As a result, slow motion is not easy to be detected. Second, slow motion is easy to be confused with background instances as they share similar characteristics that
most of the contents are unchanged in adjacent frames. Therefore, to promote the development of WTAL, a model that is not only sensitive to normal-motion actions but slow-motion actions is expected.

This paper addresses the above problems by introducing a novel framework called Slow Motion Enhanced Network (SMEN). The idea is straightforward: To make a model be sensitive to slow motion, the model should see sufficient slow motion samples during the training process. We thus propose a data mining strategy to find slow-motion-related features. SMEN consists of two modules: 1) a Mining module to generate masks for filtering out slow-motion-related features from the whole video features, and 2) a Localization module to leverage both the whole video features and minded slow-motion features to predict temporal boundaries of actions. 

Considering that only video-level labels are available under the weakly-supervised setting, the conventional mining strategy which relies on loss is not applicable. Therefore, our proposed mining module mines slow motion actions by using the prior that the action at normal speed is essentially a ``speed-up" version of its slow motion. In other words, a slow motion action can be converted, or speed-up, to normal motion through sub-sampling, so that their action characteristics become more salient and they are easier to be detected by the mining module. 
The proposed localization module then respectively processes the mined slow-motion features and the original video features with a two-branch architecture, and produces the final temporal action localization results by combining predictions of these two branches.

The contributions of this work are three-fold:
\begin{itemize}
\item[(1)] We propose a novel weakly supervised temporal action localization (WTAL) framework called Slow Motion Enhanced Network (SMEN). To the best of our knowledge, this is the first work to explore the salient slow-motion information in the WTAL task.
\item[(2)] We introduce a novel slow-motion Mining strategy, which utilizes a prior of slow-motion actions to improve their ``actionness". A two-branch localization network is proposed to handle slow- and normal-motion actions simultaneously. 

\item[(3)] Comprehensive experiments conducted on two benchmark datasets, THUMOS'14 and ActivityNet v1.3, demonstrate the effectiveness of our framework for the WTAL task. Our proposed SMEN outperforms all state-of-the-art methods by a significant margin. Furthermore, our proposed framework can be easily adapted to different base networks and effectively improve their performance in the WTAL task.
\end{itemize}

\section{Related Works}
\label{sec:relared}

\textbf{Weakly-supervised Temporal Action Localization.}\quad  More and more studies draw increasing attention to WTAL due to the time-consuming and error-prone manual labeling in a fully-supervised setting. 
UntrimmedNet~\cite{wang2017untrimmednets} introduced a classification module and a selection module for predicting a classification score and selecting relevant video segments, respectively. On top of that, STPN~\cite{nguyen2018weakly} introduces a sparse loss to enforce the sparsity of selected segments. W-TALC~\cite{paul2018w}, CoLA~\cite{zhang2021cola} and FTCL~\cite{gao2022fine} employ deep metric learning to force features from different classes to get farther distance than those from the same classes. Nguyen et al.~\cite{nguyen2018weakly} and BaS-Net~\cite{lee2020background} introduce an auxiliary background class to model background activity. ACM-NET~\cite{qu2021acm} introduces a class-agnostic action-context branch to tackle the action-context confusion issue. Wang et al.~\cite{wang2021exploring} and Li et al.~\cite{li2021weakly} utilize temporal consistency of video to refine localization results especially for less discriminative action segments. 
ACN~\cite{zhai2019action} features a new coherence loss that better supervises action boundary learning and facilitates proposal regression.

TSCN uses a self-training strategy with pseudo labels produced by two stream branches to improve the performance. UGCT~\cite{yang2021uncertainty} and Huang et al~\cite{huang2022weakly} focuses on improving the quality of pseudo labels. CO2-NET~\cite{hong2021cross} utilizes a cross-modal consensus module to filter out the information redundancy in the main modality with the help of information from different perspectives of the auxiliary modality. STAR~\cite{xu2019segregated}, 3C-NET~\cite{narayan20193c}, SF-NET~\cite{ma2020sf}, SODA~\cite{zhao2021soda}, and BackTAL~\cite{yang2021background}  use additional weak information during training and obtain huge improvement.

\textbf{Slow Motion in Video.}\quad  Slow-motion action instance often occurs in sports video~\cite{8756030,7827117,9727174,8618380} in the form of replays. Many existing work \cite{wang2004generic,chen2015novel,javed2016efficient,kiani2012effective} focus on detecting slow motion in sports by playback speed classification. However, these works usually employ a specific domain analysis. Differently, SpeedNet~\cite{benaim2020speednet} works on any videos and proposes a novel network to automatically predict the “speediness” of moving objects in videos, \textit{i.e.}, whether they are at or slower than their “natural” speed. To the best of our knowledge, most existing WTAL works do not consider that most videos contain action instances with different movements and handle all input videos with a fixed temporal scale. 
In addition, Zhu et al.~\cite{zhu2017bidirectional} also observed that the speed of motion varies constantly. They proposed a Multirate Visual Recurrent Model (MVRM) by encoding frames of a video clip with different intervals and apply it in self-supervised learning in video domain.
Unlike previous WTAL works, we introduce a data mining module to explore slow-motion information in a video and design a two-branch network for learning from original video features and the enhanced slow-motion features.

\textbf{Masking Mechanism.} \quad Most WTAL methods tend to focus on the most discriminative action segments but ignore trivial action segments, e.g., the beginning or the end of an action, which results in incomplete action localization. Therefore, masking mechanism is proposed to highlight less discriminative segments. For example, Hide-and-Seek~\cite{singh2017hide} proposes to randomly erase input segments during training, which can force the model to discover less discriminative segments. CleanNet~\cite{liu2019weakly} utilizes mask mechanism to help improve boundary regression. And more sophisticated masking mechanisms are used in later work such as Zhong et al.~\cite{zhong2018step}, ASSG~\cite{zhang2019adversarial} and A2CL-PT~\cite{min2020adversarial}. 

Although our proposed Mining module also employs a masking mechanism, it is distinguished from others as it specifically focuses on mining slow-motion-related features. Furthermore, the proposed Mining module works in a novel way by utilizing the prior that the normal-motion action can be treated as the accelerated version of the slow-motion action. Thus we can convert a slow-motion action to the corresponding normal-motion action by sub-sampling without using any additional annotations or loss feedback.

\begin{figure*}[t]
  \centering
   \includegraphics[width=\textwidth]{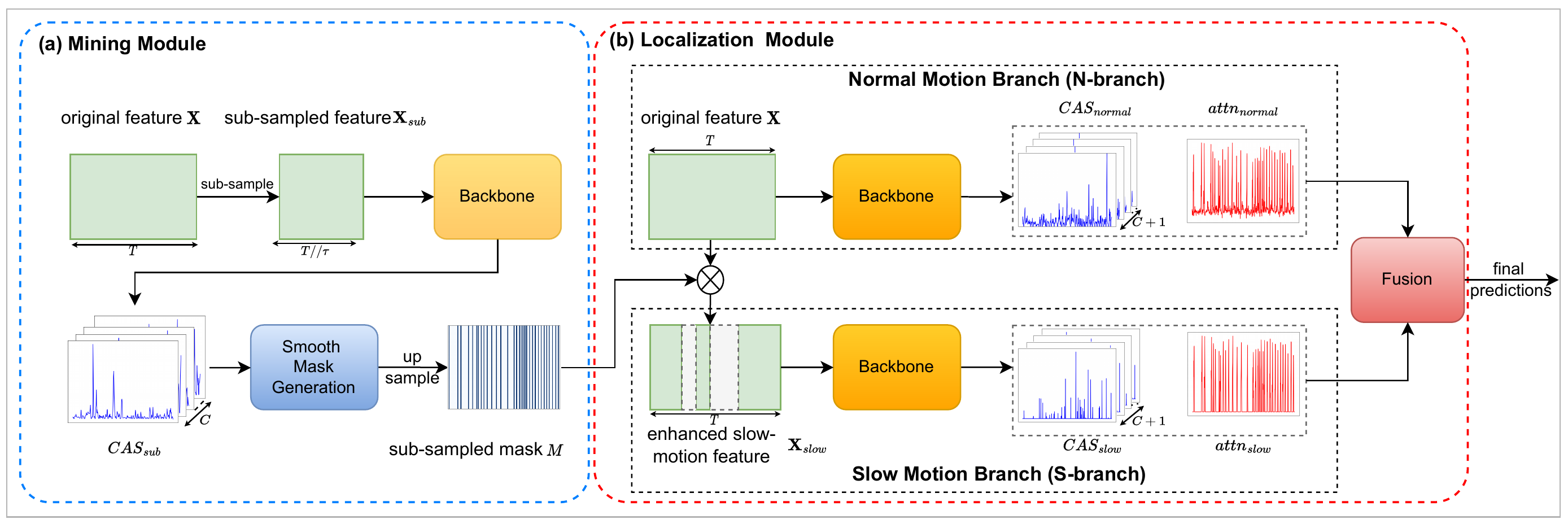}
   \caption{Overview of our proposed Slow Motion Enhanced Network (SMEN). Our proposed SMEN first extracts the video feature $\mathbf{X}$ from the original videos. (a) The Mining module applies sub-sampling operation on the original video feature $\mathbf{X}$ and uses the CAS generation backbone to generate $CAS_{sub}$, which is then used to produce the mask to generate the enhanced slow-motion feature $\mathbf{X}_{slow}$. (b) The Localization module consists of two branches with one (\textit{i.e.}, the Normal Motion Centric Branch) taking the original video feature $\mathbf{X}$ as the input while the other one (\textit{i.e.}, the Slow Motion Centric Branch) taking the enhanced slow-motion feature $\mathbf{X}_{slow}$ as the input. The CAS and the attention weights (\textit{i.e.}, $CAS_{normal}$, $attn_{normal}$, $CAS_{slow}$, $attn_{slow}$) generated from these two branches are combined via Fusion module to output the final predictions.}
   \label{fig:overview}
\end{figure*}

\section{Methodology}
\label{sec:method}

In this section, we first introduce basic notations and preliminaries in Section~\ref{subsec:3_1}. In Section~\ref{subsec:3_2}, we briefly review ACM-NET~\cite{qu2021acm}, which is used as the CAS generation backbone in our Slow Motion Enhanced Network (SMEN). We then introduce the proposed SMEN in Section~\ref{subsec:3_3}. 

\subsection{Notations and Preliminaries}
\label{subsec:3_1}

Given an untrimmed video $V$, we first divide it into non-overlapping 16-frame segments $V=\left\{v_{i}\in\mathds{R}^{16\times{H}\times{W}\times{3}}\right\}^{T}_{i=1}$ as in previous methods \cite{paul2018w,nguyen2018weakly,lee2020background}, where $T$ denotes the total number of segments. Each segment $v_{i}$ is then fed into a pre-trained feature extraction network (e.g. I3D~\cite{carreira2017quo}) to generate a $d$ dimension feature vector $x_{i} \in\mathds{R}^{d}$, and feature vectors of $T$ segments are stacked together to form a feature sequence $\mathbf{X}=\left[x_{1},x_{2},\cdots,x_{T}\right]^{\top}\in\mathds{R}^{T\times{d}}$ as the video representation. Each video has a ground-truth video-level action class label, \textit{i.e.}, a multi-hot vector $\mathbf{Y}=\left[y_{1},y_{2},\cdots,y_{C},y_{C+1}\right]^{\top}$, where $C$ is the number of action classes. $y_{c}=1,c\in[1,2,\cdots,C]$ indicates that the $c$-th action happens in the input video and $y_{C+1}=1$ indicates that the input video contains non-action background class.

\subsection{Review of ACM-NET}
\label{subsec:3_2}

Our proposed SMEN uses ACM-NET~\cite{qu2021acm} as our CAS generation backbone to produce CAS in order to obtain the sub-sampled mask and generate the temporal action localization results. Note that our proposed SMEN can use any CAS generation methods, and we take ACM-NET as an example in this work. We also evaluate our proposed SMEN on different CAS generation backbones in Section~\ref{sec:experiment}.

\textbf{ACM-NET}~\cite{qu2021acm} first utilizes a classification branch to generate the initial $CAS$ (Eq.~\ref{eq:CAS}). Then, a three-branch class-agnostic attention module is used to generate three sets of attention weights $A$ for discriminating \textit{action instance}, \textit{action context}, and \textit{non-action background}, respectively (Eq.~\ref{eq:att}).

\begin{equation}
    CAS = \Phi_{cls}(X)
\label{eq:CAS}
\end{equation}

\begin{equation}
    A = \Phi_{attn}(X)
\label{eq:att}
\end{equation}

\noindent where $CAS\in \mathds{R}^{T\times{(C+1)}}$ denotes the classification logit of each action class over time, $A = \left\{(attn_{ins}(t), attn_{con}(t), attn_{bac}(t))\right\}_{t=1}^{T} \in \mathds{R}^{T\times{3}}$ indicates the likelihood of $t$-th snippet being an  \textit{action instance}, an \textit{action context}, and a \textit{non-action background}, respectively.

Based on the attention weights generated by these three branches, ACM-NET constructs three sets of CAS as follow: 

\begin{equation}
    CAS_{*} = attn_{*} \times CAS, * = \left\{ins, con, bac\right\}
    \label{eq:CAS*}
\end{equation}

\noindent which are then aggregated to compute video-level classification scores for \textit{action instance}, \textit{action context}, and \textit{non-action background}.

These video-level classification scores are supervised by the pre-defined labels $\mathbf{Y}_{ins}=[y_{c}=1,y_{C+1}=0]$, $\mathbf{Y}_{con}=[y_{c}=1,y_{C+1}=1]$ and $\mathbf{Y}_{bac}=[y_{c}=0,y_{C+1}=1]$ for action instance, action context, and non-action background, respectively.

To train the ACM-NET~\cite{qu2021acm}, three binary cross-entropy loss are used in the objective functions for these three branches, which are denoted as $L_{ins}$, $L_{con}$ and $L_{bac}$, respectively. The whole ACM-NET is trained by jointly minimizing the overall loss function as follow:
\begin{equation}
    L_{acm-net}=L_{cls}+L_{add}
    \label{eq:l_acm1}
\end{equation}
\begin{equation}
    L_{cls}=L_{ins}+L_{con}+L_{bac}
    \label{eq:L_acm2}
\end{equation}
\begin{equation}
    L_{add}=\lambda_{1}L_{gui}+\lambda_{2}L_{feat}+\lambda_{3}L_{spa}
    \label{eq:L_acm3}
\end{equation}
where $L_{gui}$, $L_{feat}$ and $L_{spa}$ are the attention guide loss, the action feature separation loss and the sparse attention loss, respectively. We kindly refer readers to \cite{qu2021acm}  for more details about ACM-Net.

\subsection{Slow Motion Enhanced Network}
\label{subsec:3_3}

As shown in Fig.~\ref{fig:overview}, our Slow Motion Enhanced Network (SMEN) consists of a Mining module and a Localization module. The Mining module takes the original video feature as input and outputs masks used to filter slow-motion-related features. The Localization module takes both the mined slow-motion features and the original video feature to produce temporal action localization results. We will introduce the details in the following sections.

\subsubsection{Slow-motion Mining}

The proposed Mining module aims to mine slow-motion-related features. However, slow-motion actions do not have characteristics as salient as normal motions. Therefore, we first improve the ``actionness" of slow-motion actions by sub-sampling the original video feature $\mathbf{X}\in\mathds{R}^{T\times{d}}$ with ratio $\tau$, which results in a speed-up video features $\mathbf{X}_{sub}\in\mathds{R}^{(T//\tau)\times{d}}$. 

Specifically, we sample one snippet for every $\tau$ snippets from the original video feature $\mathbf{X}$ to produce the sub-sampled feature $\mathbf{X}_{sub}$. After sub-sampling, the characteristics of slow motion are enhanced, so that become easier to be detected.

We use ACM-NET as a backbone of the Mining module which generates a mask $M$ to localize slow-motion features within sub-sampled feature $\mathbf{X}_{sub}$. The CAS predicted by the backbone network are used for mask generation.

To obtain high-quality masks, we apply a smooth mask generation mechanism on the predicted CAS. In the smooth mask generation mechanism, we use the maximum values of the CAS across all action classes to represent the action activations, which are then normalized by using the Min-Max Normalization, i.e., $M^{norm}\in\mathds{R}^{(T//\tau)\times{d}}$.

Furthermore, inspired by \cite{li2021pseudo}, we use the Coefficient of Variation Smoothing to smooth the normalized action activations $M^{norm}$ based on the variation in temporal domain to remove the short temporal segments which are usually considered as noise. Specifically, we define the Coefficient of Variation $c_{v}$ as in Eq.~\ref{eq:cv}: 
\begin{equation}
    c_v=\frac{\sqrt{\mathds{D}\left(M^{norm}\right)}}{\mathds{E}\left(M^{norm}\right)}
    \label{eq:cv}
\end{equation}
where $\mathds{D}(\cdot)$ and $\mathds{E}(\cdot)$ are the functions to calculate the deviation and the mean, respectively.
We then use the Coefficient of Variation $c_v$ together with a scale factor $s$ to smooth the normalized action activations $M^{norm}$ as follows:
\begin{equation}
    M^{smooth}_{i}=\left(M_{i}^{norm}\right)^\alpha
    \label{eq:cvs}
\end{equation}
\begin{equation}
    \alpha=1-s\times c_{v}
    \label{eq:alpha}
\end{equation}
where $M^{smooth}_{i}$ and $M_{i}^{norm}$ are the $i$-th element in the smoothed action activations $M^{smooth}\in\mathds{R}^{(T//\tau)\times{d}}$ and the normalized action activations $M^{norm}$.
Then we apply a binary function to generate the mask $M$ by using a pre-defined threshold $\theta$ as following:
\begin{equation}
    M_{i}= \begin{cases} 1, &M^{smooth}_{i} \ge \theta \\ 0, &M^{smooth}_{i} < \theta \\ \end{cases} 
    \label{eq:mask2}
\end{equation}
The mask $M$ is then up-sampled to the temporal length of the original video feature $\mathbf{X}$ by nearest neighbor interpolation, finally get mask $M\in\mathds{R}^{T\times{d}}$. 

As the mask is produced based on the high action activation values generated by using the sub-sampled input feature $\mathbf{X}_{sub}$, it can be used to filter the slow-motion feature from the original video feature $\mathbf{X}$.

\subsubsection{Temporal Localization}

Once slow-motion-related features are obtained, they can be used as complementary for original video features for action localization. Precisely, the Localization module consists of two branches, a Normal-motion branch (N-branch) and a Slow-motion branch (S-branch), which both are built upon our CAS generation backbone (\textit{i.e.}, ACM-NET~\cite{qu2021acm}). Similar to ACM-NET~\cite{qu2021acm}, the N-branch takes the original video feature $\mathbf{X}$ as input and produces the CAS and attention weights. While the S-branch receives the enhanced slow-motion feature $\mathbf{X}_{slow}$ as input which are generated by filtering the original video feature with the mask produced by the Mining module. We then apply max pooling operation to combine the CAS and the attention weights generated from these two branches and use the same post-processing method and inference method as in~\cite{qu2021acm} to produce the final weakly supervised temporal action localization results. 

\subsubsection{Training Details}

For demonstration, our proposed SMEN framework uses the ACM-NET as the backbone. We use Eq.~\ref{eq:l_acm1}-\ref{eq:L_acm3} as loss functions to optimize the Mining module and Localization module separately. Note that the Mining module is first trained and then fixed to generate mask $M$.
Besides, as the Localization module consists of two branches, there are two action feature separation loss $L_{feat}^{normal}$ and $L_{feat}^{slow}$. We combine them as follows:

\begin{equation}
    L_{feat}^{fuse} = (1-\beta)L_{feat}^{normal} + \beta L_{feat}^{slow}
    \label{eq:fuse_loss}
\end{equation}
In this work, we empirically set the hyper-parameter $\beta=0.5$. During the inference stage, the Mining module is discarded as the Slow-motion branch of the Localization module is well trained to focus on the salient slow-motion features. Both branches in the Localization module take the original video feature $\mathbf{X}$ as the input.

\begin{table*}
  \centering

   \caption{Temporal localization performance comparison with state-of-the-art methods on the THUMOS'14 test set~\cite{idrees2017thumos}. Note that † represents methods that utilize external supervision information besides from video labels.}
  \resizebox{\textwidth}{!}{
    \begin{tabular}{ccccccccccccc}
    \toprule
    \multirow{2}[4]{*}{Supervision} & \multirow{2}[4]{*}{Year} & \multirow{2}[4]{*}{Methods} & \multicolumn{7}{c}{{mAP@t-IoU(\%)}} & \multirow{2}[4]{*}{AVG[0.1-0.5]} & \multirow{2}[4]{*}{AVG[0.3-0.7]} & \multirow{2}[4]{*}{AVG[0.1-0.7]} \\
\cmidrule{4-10}          &       &       & 0.10  & 0.20  & 0.30  & 0.40  & 0.50  & 0.60  & 0.70  &       &       &  \\
    \midrule
    \multirow{5}[2]{*}{Full} & 2017  & SSN~\cite{zhao2017temporal}   & 66.0  & 59.4  & 51.9  & 41.0  & 29.8  & -     & -     & 49.6  & -     & - \\
          & 2018  & BSN~\cite{lin2018bsn}   & -     & -     & 53.5  & 45.0  & 36.9  & 28.4  & 20.0  & -     & 36.8  & - \\
          & 2019  & G-TAD~\cite{xu2020g}  & -  & -  & 54.5  & 47.6  & 40.2  & 30.8     & 23.4     & -  & 39.3     & - \\
          & 2019  & GTAN~\cite{long2019gaussian}  & \textbf{69.1}  & \textbf{63.7}  & 57.8  & 47.2  & 38.8  & -     & -     & 55.3  & -     & - \\
          & 2021  & BSN++~\cite{su2020bsn++} & -     & -     & \textbf{59.9}  & \textbf{49.5}  & \textbf{41.3}  & \textbf{31.9}  & \textbf{22.8}  & -     & 41.1      & - \\
    \midrule
    \multirow{4}[2]{*}{Weak†} & 2019  & 3C-NET~\cite{narayan20193c} & 59.1  & 53.5  & 44.2  & 34.1  & 26.6  & -     & 8.1   & 43.5  & -     & - \\
          & 2020  & SF-NET~\cite{ma2020sf} & 71.0  & 63.4  & 53.2  & 40.7  & 29.3  & 18.4  & 8.6   & 51.5  & 30.2  & 40.8 \\
          & 2021  & SODA~\cite{zhao2021soda} & -  & -  & 53.1  & 44.9  & 35.6  & 26.4  & 15.8   & -  & 35.2  & - \\
          & 2021  & BackTAL~\cite{yang2021background} & -  & -  & 54.4  & 45.5  & 36.3  & 26.2  & 14.7   & -  & 35.4  & - \\
          & 2021  & LACP~\cite{lee2021learning} & \textbf{75.7}  & \textbf{71.4}  & \textbf{64.6}    & \textbf{56.5}  & \textbf{45.3}  & \textbf{34.5}  & \textbf{21.8}  & \textbf{62.7}  & \textbf{44.5}  & \textbf{52.8} \\
    \midrule
    \multirow{25}[2]{*}{Weak} & 2017  & Hide-and-seek~\cite{singh2017hide} & 36.4  & 27.8  & 19.5  & 12.7  & 6.8   & -     & -     & 20.6  & -     & - \\
          & 2018  & STPN~\cite{nguyen2018weakly}  & 52.0    & 44.7  & 35.5  & 25.8  & 16.9  & 9.9   & 4.3   & 35.0  & 18.5  & 27.0 \\
          & 2018  & W-TALC~\cite{paul2018w} & 55.2  & 49.6  & 40.1  & 31.1  & 22.8  & -     & 7.6   & 39.8  & -     & - \\
          & 2018  & Zhong et al.~\cite{zhong2018step} & 45.8  & 39.0    & 31.1  & 22.5  & 15.9  & -     & -     & 30.9  & -     & - \\
          & 2018  & STAR~\cite{xu2019segregated}†  & 68.8  & 60.0  & 48.7  & 34.7  & 23.0  & -     & -     & 47.0  & -     & - \\
          & 2019  & ASSG~\cite{zhang2019adversarial} & 65.6  & 59.4  & 50.4  & 38.7  & 25.4  & 15    & 6.6   & 47.9  & 27.2  & 37.3 \\
          & 2019  & CleanNet~\cite{liu2019weakly} & -     & -     & 37.0    & 30.9  & 23.9  & 13.9  & 7.1   & -     & 22.6  & - \\
           & 2019  & ACN~\cite{zhai2019action} & -     & -     & 35.9    & 30.7  & 24.2  & 15.7  & 7.4   & -     & 22.8  & - \\

          & 2020  & BaS-Net~\cite{lee2020background} & 58.2  & 52.3  & 44.6  & 36.0    & 27.0    & 18.6  & 10.4  & 43.6  & 27.3  & 35.3 \\

          & 2020  & A2CL-PT~\cite{min2020adversarial} & 61.2  & 56.1  & 48.1  & 39.0    & 30.1  & 19.2  & 10.6  & 46.9  & 29.4  & 37.8 \\
          & 2020  & TSCN~\cite{zhai2020two}  & 63.4  & 57.6  & 47.8  & 37.7  & 28.7  & 19.4  & 10.2  & 47.0  & 28.8  & 37.8 \\
          & 2021  & Wang et al.~\cite{wang2021exploring} & 66.1  & 60.0  & 52.3  & 43.2    & 32.9    & -  & -  & 50.9  & -  & - \\
          & 2021  & Li et al.~\cite{li2021weakly} & 67.8  & 61.9  & 54.1  & 43.7    & 32.7    & 22.1  & 12.3  & 52.0  & 33.0  & 42.1 \\
          & 2021  & ASL~\cite{ma2021weakly} & 67.0  & -  & 51.8  & -    & 31.1    & -  & 11.4  & -  & -  & - \\
          & 2021  & CoLA~\cite{zhang2021cola}  & 66.2  & 59.5  & 51.5  & 41.9  & 32.2  & 22    & 13.1  & 50.3  & 32.1  & 40.9 \\
          & 2021  & UGCT~\cite{yang2021uncertainty}  & 69.2  & 62.9  & 55.5  & 46.5  & 35.9  & 23.8  & 11.4  & 54.0    & 34.6  & 43.6 \\
          & 2021  & CO2-Net~\cite{hong2021cross} & 70.1  & 63.6  & 54.5  & 45.7  & \textbf{38.3} & \textbf{26.4} & \textbf{13.4}  & 54.0    & 35.7  & 44.6 \\
          & 2021  & ACGNet~\cite{yang2021acgnet}  & 68.1  & 62.6  & 53.1  & 44.6  & 34.7  & 22.6  & 12.0  & 52.6    & 33.4  & 42.5 \\
          & 2022  & FTCL~\cite{gao2022fine}  & 69.6  & 63.4  & 55.2  & 45.2  & 35.6  & 23.7  & 12.2  & 53.8    & 34.4  & 43.6 \\
          & 2022  & Huang et al.~\cite{huang2022weakly}  & \underline{71.3}  & 65.3  & 55.8  & \underline{47.5}  & \underline{38.2}  & \underline{25.4}  & 12.5  & \underline{55.6}    & \underline{35.9}  & \underline{45.1} \\
          & 2022  & ASM-Loc~\cite{he2022asm}  & 71.2  & \underline{65.5}  & \underline{57.1}  & 46.8  & 36.6  & 25.2  & \textbf{13.4}  & 55.4    & 35.8  & \underline{45.1} \\

        \cdashline{2-13}[1pt/1pt]
          & 2021  & ACM-NET~\cite{qu2021acm} & 68.9  & 62.7  & 55.0    & 44.6  & 34.6  & 21.8  & 10.8  & 53.2  & 33.4  & 42.6 \\
          & -     & SMEN (Ours) & \textbf{74.0} & \textbf{68.5} & \textbf{60.1} & \textbf{49.4} & 36.9  & 23.6  & \underline{12.9}  & \textbf{57.8} & \textbf{36.6} & \textbf{46.5} \\
    \bottomrule
    \end{tabular}}
  \label{tab:thumos_result}
\end{table*}

\section{Experiment}
\label{sec:experiment}

\subsection{Datasets and Evaluation Metrics}
\label{secsec:Dataset}

\subsubsection{Datasets.} \quad We perform extensive experiments on two temporal action localization benchmark datasets THUMOS'14~\cite{idrees2017thumos} and ActivityNet1.3~\cite{caba2015activitynet}.

\textbf{THMOUS14}~\cite{idrees2017thumos} contains 200 and 213 untrimmed videos for validation and testing sets, respectively, and the action instances are annotated with precise temporal action boundaries and action classes from 20 different action categories. On average, each video contains 15.4 action instances, and more than 60\% videos included slow motion. In addition, the video length varies from a few seconds to more than one hour, which makes it very challenging, especially for weakly-supervised temporal action localization. Following previous works~\cite{lee2020background,qu2021acm}, we use the videos in the validation and the testing sets for training and testing, respectively.

\textbf{ActivityNet1.3}~\cite{caba2015activitynet} contains 19,994 untrimmed videos with 10,024, 4,926, and 5,044 videos for training, validation, and testing sets, respectively. All action instances in the training and validation sets are labeled from 200 different action categories. On average, each video contains 1.6 action instances.

\textbf{HACS}~\cite{zhao2019hacs} dataset contains 50k videos spanning 200 classes, where the training/validation/testing set consists of 38k/6k/6k videos, respectively. Compared with existing benchmarks, HACS contains large-scale videos and action instances, serving as a more realistic and challenging benchmark.

\subsubsection{Evaluation Metrics.}\quad We use the mean Average Precision (mAP) with different temporal Intersection over Union (t-IoU) thresholds to evaluate our weakly-supervised temporal action localization performance, which denotes as mAP@t-IoU. Specifically, the t-IoU thresholds used to calculate the average mAP is [0.1:0.1:0.7] for THUMOS'14~\cite{idrees2017thumos}, [0.5:0.05:0.95] for ActivityNet v1.3~\cite{caba2015activitynet} and HACS~\cite{zhao2019hacs}.

\subsection{Implementation Details}
We use the I3D network~\cite{carreira2017quo}, which is pre-trained based on the Kinetics dataset~\cite{kay2017kinetics}, as our feature extractor to extract both the RGB and optical flow features. The optical flow maps are generated by using the TVL1 algorithm~\cite{zach2007duality}. Instead of extracting the features for each frame, we divide the video into a set of non-overlapping 16-frame segments and extract the RGB and flow features for each segment.

During the training process, the batch size is set to be 16, 512, and 64 for the THUMOS'14 dataset~\cite{idrees2017thumos}, the ActivityNet v1.3 dataset~\cite{caba2015activitynet}, and the HACS dataset~\cite{zhao2019hacs}, respectively. We train the proposed Mining module for 1500 iterations on the THUMOS'14 and the ActivityNet v1.3 datasets and 10 iterations on the HACS dataset. For the optimizer, we choose the Adam optimizer~\cite{kingma2014adam} for all three datasets.
The learning rate is set to be $5\times10^{-5}$, $1\times10^{-4}$, and $1\times10^{-4}$, respectively. We increased the learning rate to ten times its value to train the Localization module. The hyperparameter $r$, $\theta$ and $s$ are set to be 4, 0.4 and 0.3 for all datasets.

We kindly refer readers to \cite{qu2021acm} for other parameters of backbones. All the experiments are conducted with PyTorch~\cite{paszke2019pytorch} on a single GTX 2080Ti GPU.

\begin{table}[htbp]
  \centering
  \caption{Temporal localization performance comparison with state-of-the-art methods on the Activity-Net v1.3~\cite{caba2015activitynet} validation set. Note that † represents methods that utilize external supervision information besides from video labels.}
    \begin{tabular}{cccccc}
    \toprule
    \multirow{2}[4]{*}{Supervision} & \multirow{2}[4]{*}{Methods} & \multicolumn{3}{c}{mAP@t-IoU(\%)} & \multirow{2}[4]{*}{AVG} \\
\cmidrule{3-5}          &       & 0.50  & 0.75  & 0.95  &  \\
    \midrule
    \multirow{3}[2]{*}{Full} & SSN~\cite{zhao2017temporal}   & 39.1  & 23.5  & 5.5   & 24.0 \\
          & BSN~\cite{lin2018bsn}   & 46.5  & 30.0    & 8.0     & 30.0 \\
          & BSN++~\cite{su2020bsn++} & \textbf{51.3} & \textbf{35.7} & \textbf{8.3} & \textbf{34.9} \\
    \midrule
    \multirow{15}[2]{*}{Weak} & STPN~\cite{nguyen2018weakly}  & 29.3  & 16.9  & 2.6   & - \\
          & STAR~\cite{xu2019segregated}†  & 31.1  & 18.8  & 4.7   & - \\
          & ASSG~\cite{zhang2019adversarial}  & 32.3  & 20.1  & 4.0     & - \\
          & TSM~\cite{yu2019temporal}   & 30.3  & 19.0    & 4.5   & - \\
          & BaS-Net~\cite{lee2020background} & 34.5  & 22.5  & 4.9   & 22.2 \\
          & TSCN~\cite{zhai2020two}  & 35.3  & 21.4  & 5.3   & 21.7 \\
          & ACSNet~\cite{liu2021acsnet}  & 36.3  & 24.2  & 5.8   & 23.9 \\
          & Wang et al.~\cite{wang2021exploring} & 37.1  & 24.1  & 5.8     & 24.1 \\
          & UGCT~\cite{yang2021uncertainty}  & 39.1  & 22.4  & 5.8   & 23.8 \\
          & FAC-Net~\cite{huang2021foreground} & 37.6  & 24.2  & 6.0     & 24.0 \\
          & FTCL~\cite{gao2022fine} & 40.0  & 24.3  & \underline{6.4}     & 24.8 \\
          & Huang et al.~\cite{huang2022weakly} & 40.6  & 24.6  & 5.9     & 25.0 \\
          & ASM-Loc~\cite{he2022asm} & \underline{41.0}  & 24.9  & 6.2     & 25.1 \\
          & Li et al.~\cite{li2021weakly} & 40.9  & \textbf{25.7}  & 5.6     & \underline{25.6} \\
          \cdashline{2-6}[1pt/1pt]
          & ACM-NET~\cite{qu2021acm} & 40.1  & 24.2  & 6.2   & 24.6 \\
          & SMEN (Ours) & \textbf{41.7} & \underline{25.6} & \textbf{6.6} & \textbf{26.0} \\
    \bottomrule
    \end{tabular}
  \label{tab:A13_result}%
\end{table}%

\begin{table*}[htbp]
  \centering
  \caption{Temporal localization performance comparison between our proposed method and the state-of-the-art methods on the HACS~\cite{zhao2019hacs} validation set. Note that + denotes the methods using fully supervision, † represents the methods that utilize external supervision information in addition to the video labels. \# denotes the results are based on the implementation in BackTAL~\cite{yang2021background}, the other results are from our implementation.}

    \begin{tabular}{cccccccccccc}
    \toprule
    \multirow{2}[4]{*}{Methods} & \multicolumn{10}{c}{mAP@tIoU}                                                 & \multirow{2}[4]{*}{AVG[0.5-0.95]} \\
\cmidrule{2-11}          & 0.50  & 0.55  & 0.60  & 0.65  & 0.70  & 0.75  & 0.80  & 0.85  & 0.90  & 0.95  &  \\
    \midrule
    SSN+\#~\cite{zhao2017temporal}   & 28.8  & -     & -     & -     & -     & 18.8  & -     & -     & -     & 5.3   & 19.0  \\
    BaS-NET\#~\cite{lee2020background} & 30.6  & 27.7  & 25.1  & 22.6  & 20.0  & 17.4  & 14.8  & 12.0  & 9.2  & 5.7   & 18.5  \\
    BackTAL†\#~\cite{yang2021background} & 31.5  & 29.1  & 26.8  & 24.5  & 22.0  & 19.5  & 17.0  & 14.2  & 10.8  & 4.7   & 20.0  \\
    \midrule
    ACM-NET~\cite{qu2021acm} & 25.4  & 23.2  & 21.0  & 18.6  & 16.5  & 14.4  & 12.5  & 10.1  & 7.5   & 4.7   & 15.4  \\
    SMEN(w/ ACM-NET) & \textbf{27.6} & \textbf{25.2} & \textbf{23.1} & \textbf{20.9} & \textbf{18.7} & \textbf{16.4} & \textbf{14.1} & \textbf{11.4} & \textbf{8.5} & \textbf{5.4} & \textbf{17.1} \\
    \midrule
    BaS-NET~\cite{lee2020background} & 30.8  & 28.0  & 25.7  & 23.5  & 21.0  & 18.7  & 16.2  & 13.5  & 10.5  & \textbf{6.4} & 19.4  \\
    SMEN(w/ BaS-NET) & \textbf{32.1} & \textbf{29.4} & \textbf{26.9} & \textbf{24.3} & \textbf{21.8} & \textbf{19.4} & \textbf{16.7} & \textbf{13.9} & \textbf{10.5} & 5.9   & \textbf{20.1} \\
    \bottomrule
    \end{tabular}%

  \label{tab:hacs}%
\end{table*}%

\subsection{Comparison with the State-of-the-arts Methods}

We compare our proposed network with existing fully supervised and weakly supervised temporal action localization methods on two benchmark datasets THUMOS'14~\cite{idrees2017thumos}, ActivityNet v1.3~\cite{caba2015activitynet} and HACS~\cite{zhao2019hacs}.

\subsubsection{Results on the THUMOS'14 Dataset}\quad 
We report the mAP results on the THUMOS'14 dataset~\cite{idrees2017thumos} in Table~\ref{tab:thumos_result}. We observe that our proposed SMEN can achieve the average mAP of 46.5\%, which is 3.9\% higher than that of the baseline method ACM-NET~\cite{qu2021acm}. We believe this improvement is mainly brought by the salient slow-motion information exploitation in our proposed modules. 
We also observe that our proposed SMEN can achieve the best \textit{Average mAP} results and outperforms baseline models ACM-NET~\cite{qu2021acm}  by 7\%, 9\%, 9\%, 11\%, 7\%, 8\%,
19\% at t-IoU=[0.1,:0.1,0.7] in terms of \textit{relative improvement}, indicating that our proposed SMEN have consistent improvement across all t-IoU thresholds.  
It is worthy to note that the methods like STAR~\cite{xu2019segregated}, 3C-NET~\cite{narayan20193c},  SF-NET~\cite{ma2020sf}, SODA~\cite{zhao2021soda}, and BackTAL~\cite{yang2021background}, use additional information as supervision during training instead of using the video-level annotations as the only supervision. However, our proposed SMEN outperforms all these methods in terms of all evaluation metrics. Furthermore, our proposed SMEN even outperforms the fully supervised method SSN.

\begin{table}[htbp]
  \centering
  \caption{Performance comparison of several variations for our proposed method on the THUMOS'14 test set~\cite{idrees2017thumos}. The average mAP (\%) is computed at t-IoU thresholds [0.1:0.1:0.7].``TB'' denotes Two Branches; ``MM'' denotes Mining Module; ``SSF'' denotes sub-sampled feature.}
  \resizebox{\columnwidth}{!}{
    \begin{tabular}{ccccccccccc}
    \toprule
    \multirow{2}[4]{*}{TB} & \multirow{2}[4]{*}{MM} & \multirow{2}[4]{*}{SSF} & \multicolumn{7}{c}{mAP@tIoU}                          & \multirow{2}[4]{*}{AVG[0.1-0.7]} \\
\cmidrule{4-10}          &       &       & 0.10  & 0.20  & 0.30  & 0.40  & 0.50  & 0.60  & 0.70  &  \\
    \midrule
    \XSolidBrush     & \XSolidBrush     & \XSolidBrush     & 68.9  & 62.7  & 55.0    & 44.6  & 34.6  & 21.8  & 10.8  & 42.6 \\
    \Checkmark     & \XSolidBrush     & \XSolidBrush     & 69.9 & 63.1 & 54.7 & 44.2 & 32.2 & 22.3 & 11.7 & 42.6 \\
    \Checkmark     & \XSolidBrush     & \Checkmark     & 71.4  & 64.8  & 56.0    & 45.9  & 33.9  & 21.5  & 11.0    & 43.5 \\
    \Checkmark     & \Checkmark     & \XSolidBrush     & 73.4 & 66.4 & 57.8 & 46.6 & 34.2 & 21.7 & 11.4 & 44.5 \\
    \Checkmark     & \Checkmark     & \Checkmark     & \textbf{74.0} & \textbf{68.5} & \textbf{60.1} & \textbf{49.4} & \textbf{36.9} & \textbf{23.6} & \textbf{12.9} & \textbf{46.5} \\
    \bottomrule
    \end{tabular}%
    }
  \label{tab:ablation1}%
\end{table}%

\subsubsection{Results on the Activity-Net v1.3}\quad In Table~\ref{tab:A13_result}, we report the mAPs of our proposed SMEN and compare state-of-the-art approaches at different t-IoU thresholds (\textit{i.e.}, t-IoU={0.5, 0.75, 0.95}) on the validation set of the Activity-Net v1.3 dataset. Besides, we also report the average mAP, calculated by averaging all mAPs across multiple t-IoU thresholds ranging from 0.5 to 0.95 with an interval of 0.05. We observe that our proposed SMEN outperforms all state-of-the-art WTAL approaches in terms of all evaluation metrics, including its CAS generation backbone ACM-NET. Similar to our observations on the THUMOS'14 dataset, it can be seen that the average mAP of our proposed SMEN is higher than that of some fully supervised temporal action localization approaches (\textit{e.g.}, SSN~\cite{zhao2017temporal}) and STAR~\cite{xu2019segregated}, which takes advantage of extra supervision information other than video-level labels.

\subsubsection{Results on the HACS}\quad We conduct experiments on HACS dataset with the baseline methods ACM-NET~\cite{qu2021acm} and BaS-NET~\cite{lee2020background}.
The experimental results are shown in Table~\ref{tab:hacs}. Our method can achieve the average mAP of 17.1\% and 20.1\%, which is 1.7\% and 0.7\% higher than baseline methods ACM-NET and BaS-NET, respectively. The improvements mainly come from the salient slow-motion information exploitation in our proposed modules.
It is worthy mentioning that our proposed SMEN, which uses the video-level annotation as the only supervision, outperforms the SOTA method BackTAL~\cite{yang2021background} slightly, which uses additional weak information during training. Furthermore, our proposed SMEN even outperforms the state-of-the-art fully supervised approach SSN~\cite{zhao2017temporal}.

\begin{table}[htbp]
  \centering
  \caption{Performance comparison of different $\beta$ for our proposed method on the THUEMOS'14 test set. The average mAP (\%) is computed at t-IoU thresholds [0.1:0.1:0.7].}
  \resizebox{\columnwidth}{!}{
    \begin{tabular}{ccccccccc}
    \toprule
    \multirow{2}[4]{*}{$\beta$} & \multicolumn{7}{c}{mAP@t-IoU}                          & \multirow{2}[4]{*}{AVG[0.1-0.7]} \\
\cmidrule{2-8}          & 0.10  & 0.20  & 0.30  & 0.40  & 0.50  & 0.60  & 0.70  &  \\
    \midrule
    0.00     & 71.3  & 65.0    & 56.8  & 47.0    & 35.1  & 21.8  & 11.7  & 44.1 \\
    0.25   & 72.6  & 66.8  & 58.7  & 47.7  & 35.8  & 23.3  & 12.3  & 45.3 \\
    0.50   & \textbf{74.0}    & \textbf{68.5}  & \textbf{60.1}  & \textbf{49.4}  & \textbf{36.9}  & \textbf{23.6}  & \textbf{12.9}  & \textbf{46.5} \\
    0.75   & 73.6  & 68.2  & 59.4  & 49.0    & 36.4  & 23.9  & 13.1  & 46.2 \\
    1.00     & 55.1  & 47.1  & 37.6  & 26.5  & 16.4  & 9.0     & 2.9   & 27.8 \\
    \bottomrule
    \end{tabular}%
    }
  \label{tab:beta}%
\end{table}%
\begin{table}[htbp]
  \centering
  \caption{Performance comparison between different SOTA backbones and our proposed method on the THUEMOS’14 slow-motion testing set. The average mAP (\%) is computed at t-IoU thresholds [0.1:0.1:0.7].}
  \resizebox{\columnwidth}{!}{
    \begin{tabular}{ccccccccc}
    \toprule
    \multirow{2}[4]{*}{Methods} & \multicolumn{7}{c}{mAP@tIoU}                          & \multirow{2}[4]{*}{AVG[0.1-0.7]} \\
\cmidrule{2-8}          & 0.10  & 0.20  & 0.30  & 0.40  & 0.50  & 0.60  & 0.70  &  \\
    \midrule
    BaS-NET & 43.5  & 40.4  & 35.4  & 31.5  & 22.7  & 18.7  & 14.2  & 29.5 \\
    SMEN(w/ BaS-NET) & \textbf{49.0} & \textbf{45.3} & \textbf{41.1} & \textbf{35.9} & \textbf{27.5} & \textbf{20.9} & \textbf{14.3} & \textbf{33.4} \\
    \midrule
    ASL   & 48.3  & 43.9  & 37.5  & 33.2  & 26.1  & 20.1  & 14.1  & 31.9 \\
    SMEN(w/ ASL) & \textbf{50.9} & \textbf{47.9} & \textbf{43.3} & \textbf{38.5} & \textbf{32.7} & \textbf{26.2} & \textbf{19.4} & \textbf{37.0} \\
    \midrule
    ACM-NET & 55.8  & 52.1  & 47.2  & 40.4  & 31.9  & 23.3  & 13.4  & 37.7 \\
    SMEN(w/ ACM-NET) & \textbf{58.7} & \textbf{54.5} & \textbf{48.5} & \textbf{42.6} & \textbf{35.2} & \textbf{24.8} & \textbf{16.0} & \textbf{40.0} \\
    \bottomrule
    \end{tabular}%
    }
  \label{tab:slow}%
\end{table}%

\subsection{Model Analysis}

In this section, we perform a series of experiments on the THUMOS'14 dataset~\cite{idrees2017thumos} to further demonstrate the contribution of individual components of our proposed SMEN, and provide more insights about this task. 
Unless specified, all the reported results are achieved by using ACM-NET~\cite{qu2021acm} as the CAS generation backbone.

\subsubsection{Ablation Studies}\quad Given that our proposed Localization module consists of two branches, we start from a plain two-branch baseline model, which is essentially an ensemble model of two ACM-NET models. As shown in the first and second row of Table~\ref{tab:ablation1}, the temporal action localization performance cannot be improved by simply combining two ACM-NET models, i.e., both the baseline model and the two-branch variant achieve the mAP of 42.6\%.

\textbf{Sub-sampled feature.} On top of the baseline model, we replace the input of one branch with the sub-sampled features in which the slow-motion-related features are enhanced. We can see that by using sub-sampled features, it can bring about 1\% improvement (43.5\% vs. 42.6\%), which verifies the effectiveness of the prior used in our method, i.e., normal motion actions can be treated as the speed-up version of slow motion, and the characteristics of the slow motion will become more salient after speeding up.

\textbf{Mining module.} The key of our proposed SMEN is the slow-motion mining module. In our proposed SMEN, the mining module takes the sub-sampled feature as the input during the training process and outputs a mask to be used for filtering slow-motion-related features. As shown in the third and fifth rows of Table~\ref{tab:ablation1}, the mining module significantly increases the performance from 43.5\% to 46.5\%, i.e., 3\% improvement.

It is interesting to see that even only with the original video features, the mining module can increase the average mAP by 1.9\% (44.5\% vs. 42.6\%) as shown in the second and fourth row. A possible explanation is that the mining module learns to filter out less discriminative features from the original video features for the action-instances with normal motion. This is because a pre-defined threshold (Eq.~\ref{eq:mask2}) is used to generate masks by preserving areas with high activation scores, which helps to suppress false positive predictions. 

\textbf{Balance Coefficient $\beta$.} 
We conduct an ablation study on $\beta$ and report the results in Table~\ref{tab:beta}.  A bigger $\beta$ means the action feature separation loss generated from the \textit{S}-branch produces a larger weight. When $\beta=0.5$, the model achieves the best performance.

\textbf{Performance on Slow Motion.}\quad To test the performance of the proposed method on handling slow-motion actions, a slow-motion testing set is formed by manually annotating the slow-motion action instances in THUMOS'14 testing set. Three volunteers are involved. 
This slow-motion testing set consists of 126 videos with 839 slow-motion action instances from all 20 action categories. That means there are more than 60\% videos and 25\% action instances of the THUMOS'14 testing set contain slow motion. 

We compare the WTAL performance on slow-motion action instances of three baseline method, \textit{i.e.} BaS-NET~\cite{lee2020background}, ASL~\cite{ma2021weakly} and ACM-NET~\cite{qu2021acm} and our proposed SMEN. Note that we replaced the feature corresponding to normal-motion action with $\mathbf{0} \in \mathds{R}^{d}$ during testing.
The results are reported in Table~\ref{tab:slow}. We can see that our proposed SMEN achieves the average mAP of 33.4\%, 37.0\%, 40.0\%, surpassing the baseline methods BaS-NET(29.5\%), ASL(31.9\%), ACM-NET (37.7\%) by 3.9\%, 5.1\%, 2.3\%, respectively. This margin demonstrates the superiority of our proposed SMEN in localizing slow-motion actions. All results listed in Table~\ref{tab:ablation1} and Table~\ref{tab:slow} show that our proposed mining module and the two-branch localization module can work cooperatively to enhance the model`s localization ability.

\begin{figure*}[!htbp]
\centering
\subfloat[\small \textrm{An example of action class ``SoccerPenalty''.}]{\includegraphics[width=0.95\textwidth]{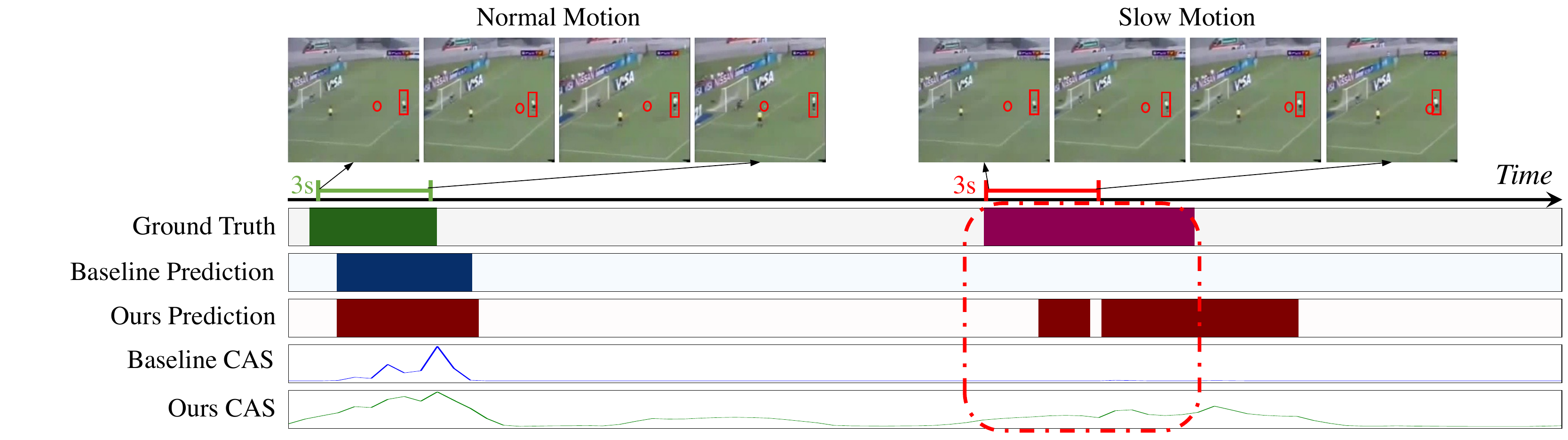}%
\label{fig:v1}}
\hfil
\subfloat[\small \textrm{An example of action class ``Blizzards''.}]{\includegraphics[width=0.95\textwidth]{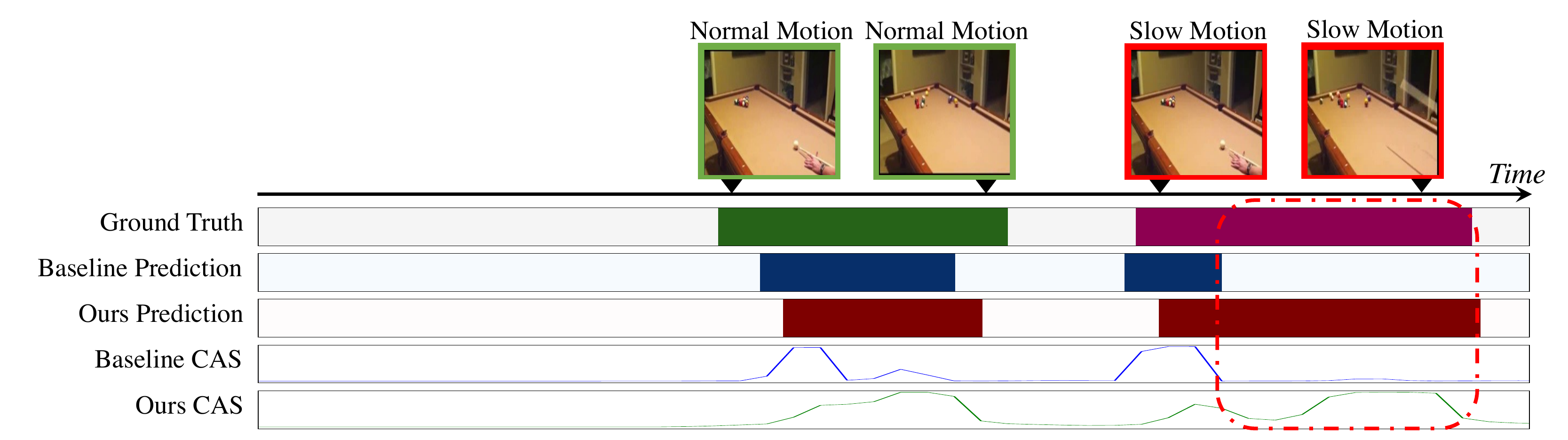}%
\label{fig:v2}}
\hfil
\subfloat[\small \textrm{An example of action class ``LongJump''.}]{\includegraphics[width=0.95\textwidth]{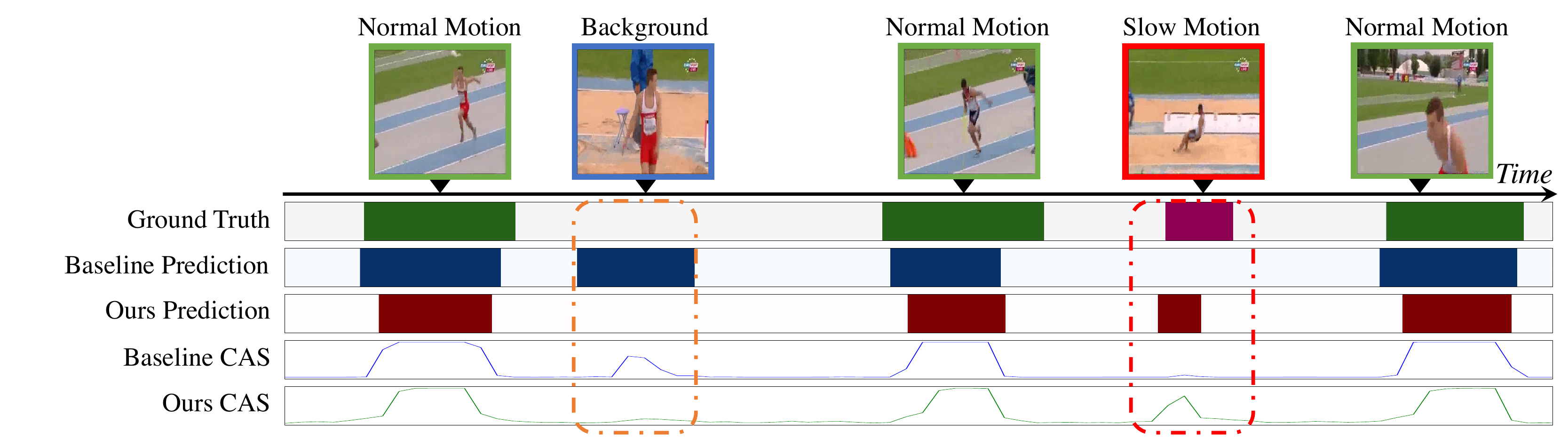}%
\label{fig:v3}}
\hfil
\subfloat[\small \textrm{An example of action class ``Diving''.}]{\includegraphics[width=0.95\textwidth]{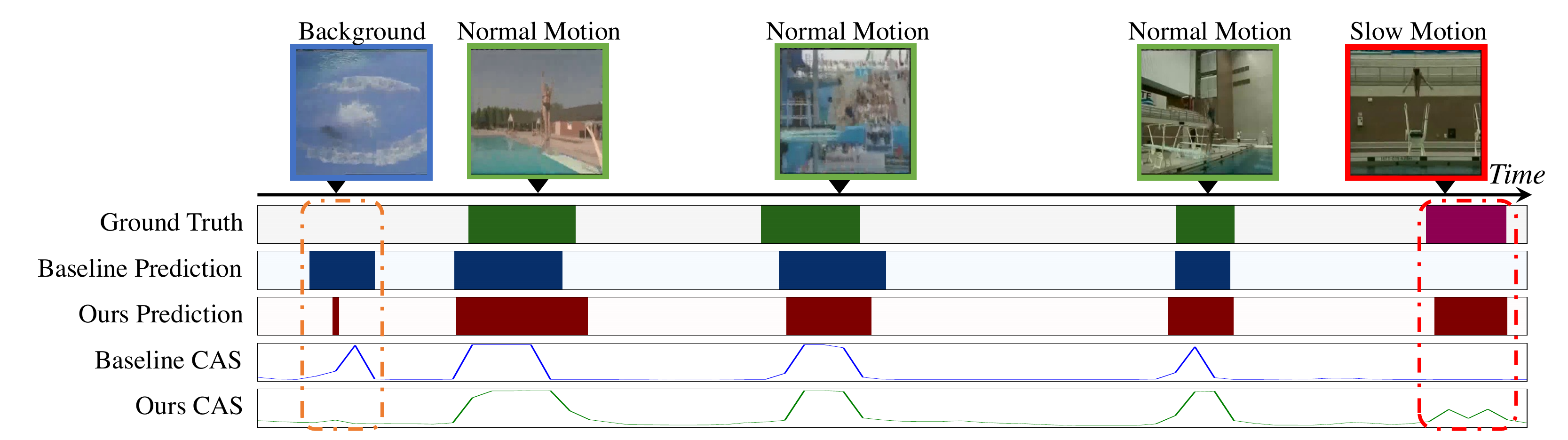}%
\label{fig:v4}}
\caption{Qualitative comparisons with baseline on THUMOS'14 dataset. The barcode is the ground-truth (GT). Specifically, the green barcode is the action instance with \textit{normal motion} while the red barcode is the action instance with \textit{slow motion}. The following line charts are CAS value of baseline model (ACM-NET~\cite{qu2021acm}) and our proposed SMEN, respectively. For beteer clarity, the frames with green bounding boxes refer to ground-truth actions with \textit{normal motion}, the frames with red bounding boxes refer to ground-truth actions with \textit{slow motion}, while those in blue refer to ground-truth \textit{backgrounds}. The red dotted boxes show that our proposed SMEN performs better in localizing slow-motion action instances. The yellow dotted box shows that SMEN can suppress the false positive instance, \textit{i.e.}, \textit{Background}$\rightarrow$\textit{Action}, incorrectly predicted by the baseline model (ACM-NET~\cite{qu2021acm}).}
\label{fig:visual}
\end{figure*}

\begin{figure*}[!htbp]
\centering
\subfloat[\small \textrm{A failure example of action class ``JavelinThrow''.}]{\includegraphics[width=0.95\textwidth]{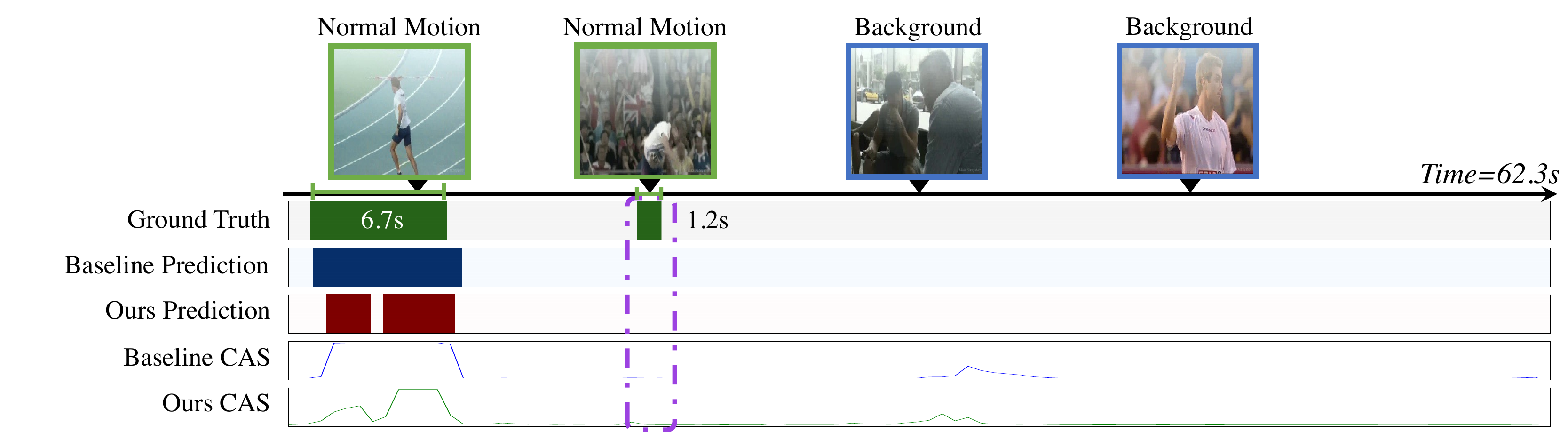}%
\label{fig:f1}}
\hfil
\subfloat[\small \textrm{A failure example of action class ``Diving''.}]{\includegraphics[width=0.95\textwidth]{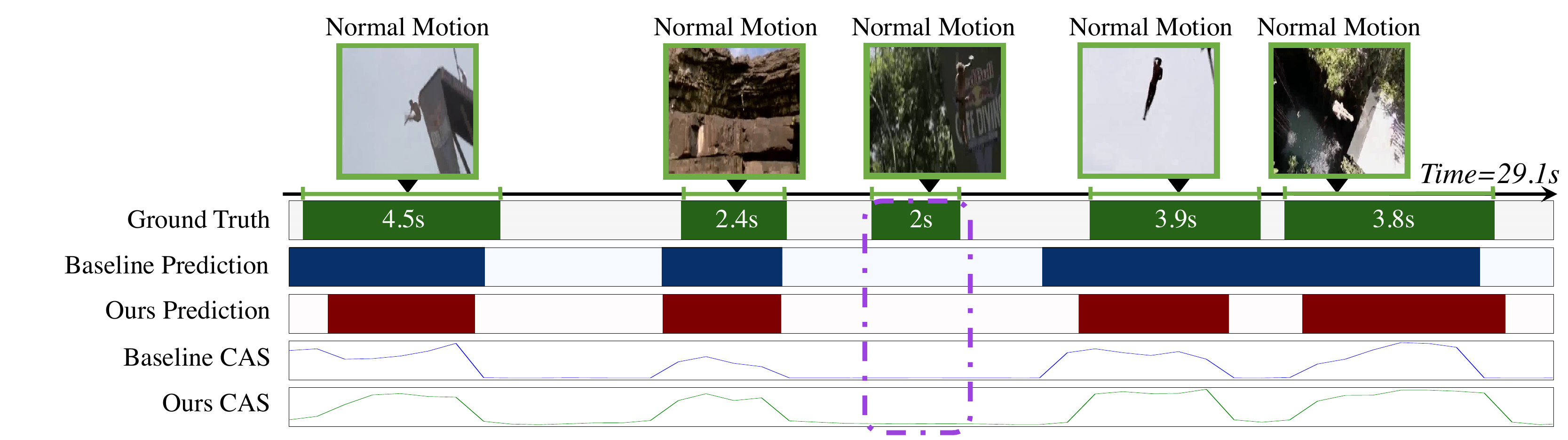}%
\label{fig:f2}}
\caption{Failure case of our proposed SMEN. The barcode is the ground-truth (GT). Specifically, the green barcode is the action instance. The following line charts are CAS value of baseline model (ACM-NET~\cite{qu2021acm}) and our proposed SMEN, respectively. For better clarity, the frames with green bounding boxes refer to ground-truth actions with \textit{normal motion}, while those in blue refer to ground-truth \textit{backgrounds}. The purple dotted boxes show that our proposed SMEN can not well localize the action instances of short duration. }
\label{fig:fail}
\end{figure*}

\begin{table}[htbp]
  \centering
  \caption{Performance comparison of several variants for our proposed method on the THUMOS'14 testing set. \textit{N}-branch denotes the results from \textit{N}ormal-branch, while \textit{S}-branch denotes the results from \textit{S}low-branch. SMEN-combo donates the variant where CAS is from the \textit{N}ormal-branch while attention values from the \textit{S}low-branch. The average mAP (\%) is computed at t-IoU thresholds [0.1:0.1:0.7].}
  \resizebox{\columnwidth}{!}{
    \begin{tabular}{ccccccccc}
    \toprule
    \multirow{2}[3]{*}{sub-sampled feature} & \multicolumn{7}{c}{mAP@tIoU}                          & \multirow{2}[3]{*}{AVG[0.1-0.7]} \\
\cmidrule{2-8}          & 0.10  & 0.20  & 0.30  & 0.40  & 0.50  & 0.60  & 0.70  &  \\
    \midrule
    N-branch & 70.8  & 63.3  & 53.2  & 41    & 28.9  & 18.1  & 9.7   & 40.7 \\
    S-branch & 0.1   & 0.1   & 0.1   & 0.1   & 0.1   & 0.0     & 0.0     & 0.1 \\
    SMEN  & \textbf{74.0}    & \textbf{68.5}  & \textbf{60.1}  & \textbf{49.4}  & \textbf{36.9}  & \textbf{23.6}  & \textbf{12.9}  & \textbf{46.5} \\
    SMEN-combo & 73.5  & 67.6  & 59.2  & 48.0  & 34.8  & 22.7  & 12.1  & 45.4  \\
    \bottomrule
    \end{tabular}%
    }
  \label{tab:SME}%
\end{table}%

\subsubsection{The Role of Each Branch in Localization Module}\quad
We additionally conduct experiments to show the role of individual branch, i.e., Normal-motion branch (N-branch) and Slow-motion branch (S-branch), of the localization module. We first \textit{directly} evaluate their performance respectively, and the results are shown in Table~\ref{tab:SME}. It is not surprisingly to see that the S-branch performs poorly on this task since this branch only ``see" slow-motion-related features during training. 

To reveal the role of the S-branch, we provide the performance of the variant that using the CAS predictions of the N-branch and the attention predictions of the S-branch, i.e., SMEN-combo in Table~\ref{tab:SME}. In other words, SMEN-combo differs from the N-branch in that it uses the attention values predicted by the S-branch to compute the final CAS (Eq.~\ref{eq:CAS*}). 
As shown in Table~\ref{tab:SME}, SMEN-combo outperforms \textit{N}-branch by 4.7\%, suggesting that the \textit{S}-branch alone is incapable of localizing a specific action, but it helps \textit{N}-branch by providing more accurate attention weights. To be more specific, originally the slow-motion actions are easily confused with action context or non-action background by the N-branch. Since the S-branch learns from the enhanced slow-motion features, it becomes more sensitive than the N-branch in discriminating action-instance, action-context, and non-action background.

\subsubsection{Generalization}\quad Our proposed SMEN framework can be easily adapted to different baseline methods. We conduct experiments by changing the CAS generation backbone with different existing networks, such as BaS-NET~\cite{lee2020background} and ASL~\cite{ma2021weakly}. The average mAPs (at t-IoU thresholds [0.1:0.1:0.9]) of our proposed SMEN with different CAS generation backbones are reported in Table~\ref{tab:ablation5}. We observe that SMEN (w/ BaS-NET) and SMEN (w/ ASL) can achieve the average mAP of 29.9\% and 34.2\%, which are 2.0\% higher than those of the baseline methods BaS-NET~\cite{lee2020background} and ASL~\cite{ma2021weakly}, respectively. This indicates that our proposed SMEN can generalize to different CAS generation methods and consistently improve their performance in the task of weakly-supervised temporal action localization.

\begin{table}[tbp]
  \centering
  \caption{Performance comparison of our proposed method when using different backbones on the THUMOS'14 test set~\cite{idrees2017thumos}. The average mAP (\%) is computed at t-IoU thresholds [0.1:0.1:0.9].}
  \resizebox{\columnwidth}{!}{
    \begin{tabular}{ccccccccccc}
    \toprule
    \multirow{2}[4]{*}{Methods} & \multicolumn{9}{c}{mAP@tIoU}                                          & \multirow{2}[4]{*}{AVG[0.1-0.9]} \\
\cmidrule{2-10}          & 0.10  & 0.20  & 0.30  & 0.40  & 0.50  & 0.60  & 0.70  & 0.80  & 0.90  &  \\
    \midrule
    BaS-NET & 58.2  & 52.3  & 44.6  & 36.0    & 27.0    & 18.6  & 10.4  & 3.9   & 0.5   & 27.9 \\
    SMEN(w/ BaS-NET) & \textbf{60.7}  & \textbf{55.3}  & \textbf{47.9}  & \textbf{38.5}  & \textbf{28.5}  & \textbf{21.4}  & \textbf{11.8}  & \textbf{4.5}   & \textbf{0.7}   & \textbf{29.9} \\
    \midrule
    ASL   & 67.0    & -     & 51.8  & -     & 31.1  & -     & 11.4  & -     & 0.7   & 32.2 \\
    SMEN(w/ ASL) & \textbf{69.0}    & \textbf{63.7}  & \textbf{54.9}  & \textbf{43.7}  & \textbf{32.9}  & \textbf{22.7}  & \textbf{14.3}  & \textbf{6.5}   & \textbf{1.0}     & \textbf{34.2} \\
    \bottomrule
    \end{tabular}%
    }
  \label{tab:ablation5}%
\end{table}%

\begin{table}[t!]
  \centering
  \caption{Performance comparison of our proposed method when using different mask generation methods on the THUMOS'14 test set~\cite{idrees2017thumos}. The average mAP (\%) is computed at t-IoU thresholds [0.1:0.1:0.7].}
  \resizebox{\columnwidth}{!}{
    \begin{tabular}{ccccccccc}
    \toprule
    \multirow{2}[3]{*}{mask generation methods} & \multicolumn{7}{c}{mAP@tIoU}                          & \multirow{2}[3]{*}{AVG[0.1-0.7]} \\
\cmidrule{2-8}          & 0.10  & 0.20  & 0.30  & 0.40  & 0.50  & 0.60  & 0.70  &  \\
    \midrule
    normal & 73.4  & 67.6  & 59.4  & 48.9  & 36.5  & 23.7  & 13.3  & 46.1 \\
    smooth & 74.0    & 68.5  & 60.1  & 49.4  & 36.9  & 23.6  & 12.9  & 46.5  \\
    \bottomrule
    \end{tabular}%
    }
  \label{tab:mask}%
\end{table}%

\subsubsection{Smooth Mask Generation}\quad  In order to verify the effectiveness of the proposed Smooth Mask Generation strategy used in the Mining module, we conduct experiments by comparing two different methods: 1) generating mask with a fixed threshold, \textit{i.e.}, \textit{normal mask}; 2) generating mask with the proposed Coefficient of Variation Smoothing mechanism which is inspired by \cite{li2021pseudo}, \textit{i.e.}, \textit{smooth mask}. Our proposed SMEN adopts the latter option.
Table~\ref{tab:mask} compares the average mAPs achieved by using these two mask generation methods on the THUMOS`14 dataset. 
We can observe that using the smooth mask generation mechanism can achieve a 0.4\% higher average mAP than using the normal mask. A possible explanation for this is that the smooth mask generation mechanism can de-noise CAS values and produce more robust masks for selecting slow-motion-related features.

\begin{table}[tbp]
  \centering
  \caption{Performance comparison of our proposed method when using different sub-sampling methods on the THUMOS14 test set~\cite{idrees2017thumos}. The average mAP (\%) is computed at t-IoU thresholds [0.1:0.1:0.7].}
  \resizebox{\columnwidth}{!}{
    \begin{tabular}{ccccccccc}
    \toprule
    \multirow{2}[3]{*}{sub-sampled methods} & \multicolumn{7}{c}{mAP@tIoU}                          & \multirow{2}[3]{*}{AVG[0.1-0.7]} \\
\cmidrule{2-8}          & 0.10  & 0.20  & 0.30  & 0.40  & 0.50  & 0.60  & 0.70  &  \\
    \midrule
    frame & 74.1  & 68.1  & 58.9  & 47.9  & 36.0    & 22.9  & 12.1  & 45.8 \\
    feature & 74.0    & 68.5  & 60.1  & 49.4  & 36.9  & 23.6  & 12.9  & 46.5  \\
    \bottomrule
    \end{tabular}%
    }
  \label{tab:sample method}%
\end{table}%

\begin{table}[htbp]
  \centering
  \caption{Performance comparison of our proposed method when using different sub-sampling rate on the THUMOS14 test set~\cite{idrees2017thumos}. The average mAP (\%) is computed at t-IoU thresholds [0.1:0.1:0.7].}
  \resizebox{\columnwidth}{!}{
    \begin{tabular}{ccccccccc}
    \toprule
    \multirow{2}[3]{*}{$\tau$} & \multicolumn{7}{c}{mAP@tIoU}                          & \multirow{2}[3]{*}{AVG[0.1-0.7]} \\
\cmidrule{2-8}          & 0.10  & 0.20  & 0.30  & 0.40  & 0.50  & 0.60  & 0.70  &  \\
    \midrule
    1     & 73.4  & 66.4  & 57.8  & 46.6  & 34.2  & 21.7  & 11.4  & 44.5 \\
    2     & 73.5  & 67.7  & 58.8  & 47.8  & 35.4  & 22.9  & 12.0    & 45.5 \\
    4     & \textbf{74.0} & \textbf{68.5} & \textbf{60.1} & \textbf{49.4} & \textbf{36.9} & \textbf{23.6} & \textbf{12.9} & \textbf{46.5} \\
    8     & 72.5  & 66.5  & 58.1  & 48.0  & 35.4  & 23.4  & 12.4  & 45.2  \\
    \bottomrule
    \end{tabular}%
    }
  \label{tab:sample rate}%
\end{table}%

\begin{table}[t]
  \centering
  \caption{Comparison between our proposed method and the baseline methods in terms of memory usage, training time and inference time on the THUMOS'14 dataset~\cite{idrees2017thumos}.}
  \resizebox{\columnwidth}{!}{
    \begin{tabular}{cccc}
    \toprule
    Methods & Memory & \multicolumn{1}{l}{traing time(s/epoch)} & \multicolumn{1}{c}{inference time(s/video)} \\
    \midrule
    BaS-NET & 3536MB & 6.64  & 0.02  \\
    SMEN(w/ BaS-NET)  & 5430MB & 11.27  & 0.03  \\
    \midrule
    ASL & 1743MB & 3.87  & 0.03  \\
    SMEN(w/ ASL)  & 2239MB & 4.64  & 0.04  \\
    \midrule
    ACM-NET & 2335MB & 4.92  & 0.07  \\
    SMEN(w/ ACM-NET)  & 3201MB & 7.93  & 0.10  \\
    \bottomrule
    \end{tabular}%
    }
  \label{tab:eff}%
\end{table}%

\subsubsection{Sub-sampling Strategy} \textbf{Feature-level vs. Frame-level.} We sub-sample on video features in our proposed framework to speed-up slow motion to normal motion. However, we can alternatively speed up the motions by sub-sampling on \textit{video frames}. Therefore, we compare the performance of using these two sub-sampling strategies. 

Table~\ref{tab:sample method} shows the average mAPs achieved by using different sub-sampling methods on the THUMOS`14 dataset. We can see that sub-sampling on video features can achieve 0.7\% higher average mAP than sub-sampling on input video frames. Apart from superior performance, sub-sampling on video features is more efficient as it does not require re-extracting video features by a pre-trained feature extractor. 

\textbf{Sub-sampling Rates.} We also investigate the influence of different sub-sampling rates for generating sub-sampled features. The results are reported in Table~\ref{tab:sample rate}. It can be seen that when the sub-sampling rate $\tau=4$, our proposed SMEN can achieve the best temporal action localization performance. A possible explanation is that a lower sub-sampling rate may not speed up the slow motion actions enough to stand out their salient characteristics, while a larger sub-sampling rate may lead to losing too much temporal information, making it hard to discriminate from background instances.

\subsubsection{Efficiency}\quad We compare the efficiency of several baseline methods and our proposed SMEN on the machine with a single GTX 2080Ti GPU. The results are shown in Table~\ref{tab:eff}. The inference time of our proposed SMEN is slightly higher than the baseline methods.

\subsection{Qualitative Results}

We visualize the localized regions and the CAS results for four actions on the THUMOS'14 dataset in Fig.~\ref{fig:visual}. Our proposed SMEN has a more informative CAS distribution compared to baseline method, thus leading to more accurate localization for slow-motion segments. 

Figure~\ref{fig:v1} depicts a typical case (``SoccerPenalty'') that the video contains two segments, one is normal motion and the other is slow motion. Specifically, the second action segment is the slow-motion replay of the first. As shown in Fig.~\ref{fig:v1}, both segments begin from preparing Soccer Penalty. After 3-second interval, the normal-motion segment turns to finish, but the slow-motion segment just turns to torch the ball. With the Mining module, the Localization module can leverage the generated mask to filter slow-motion-related features as complementary information to improve the temporal action localization results. As a result, while the backbone model fails to localize the slow-motion action segments, our proposed SMEN can accurately localize the action instance at a much slower speed. Another example in shown in Fig.~\ref{fig:v2}.

On the other hand, by introducing the Mining module, our proposed SMEN can filter out less discriminative features, thereby avoiding many false positives produced by a single branch (backbone). Fig.~\ref{fig:v3} demonstrates an example of “LongJump” action in which the backbone model incorrectly localizes a background segment (highlighted in blue box). 
This background segment belongs to the class of ``action context" which is not the target action but contains similar scenes and slow movements. Such action context segments are easily confused with actions, especially slow motion actions. Another example in shown in Fig.~\ref{fig:v4}.

By considering the slow motion, our proposed SMEN can distinguish action-instances from slow-motion-alike action-contexts, and therefore, improve the WTAL results.

\textbf{Failure Cases}. 
As shown in Fig.~\ref{fig:fail}, the baseline method and our proposed SMEN cannot well handle the action instances with very short duration well (highlighted in purple dotted boxes). A possible explanation is both the baseline and our method do not have any specific designs for such cases. We will leave this problem in our future work.


\section{Conclusion}
\label{sec:conclusion}

This paper has proposed a novel framework, SMEN, to address the problem that slow motion provides little information for the WTAL frameworks to understand the content and distinguish them from action instances. The new framework consists of two modules, a Mining module and a Localization module. The Mining module is proposed to generate slow-motion-related mask and the Localization module is designed to leverage the generated mask to select enhanced slow-motion features as complementary information to improve the temporal action localization results. Experiments conducted on three benchmarks, including THUMOS'14, ActivityNet v1.3, and HACS, have validated the state-of-the-art performance of SMEN.

\textbf{Acknowledgement:} This work is supported by the National Natural Science Foundation of China (No. 62002012) and the National Key Research and Development Project of China (No. 2018AAA0101900).

\bibliographystyle{IEEEtran}
\bibliography{ref}

\newpage

\begin{IEEEbiography}
[{\includegraphics[width=1in,height=1.25in,clip,keepaspectratio]{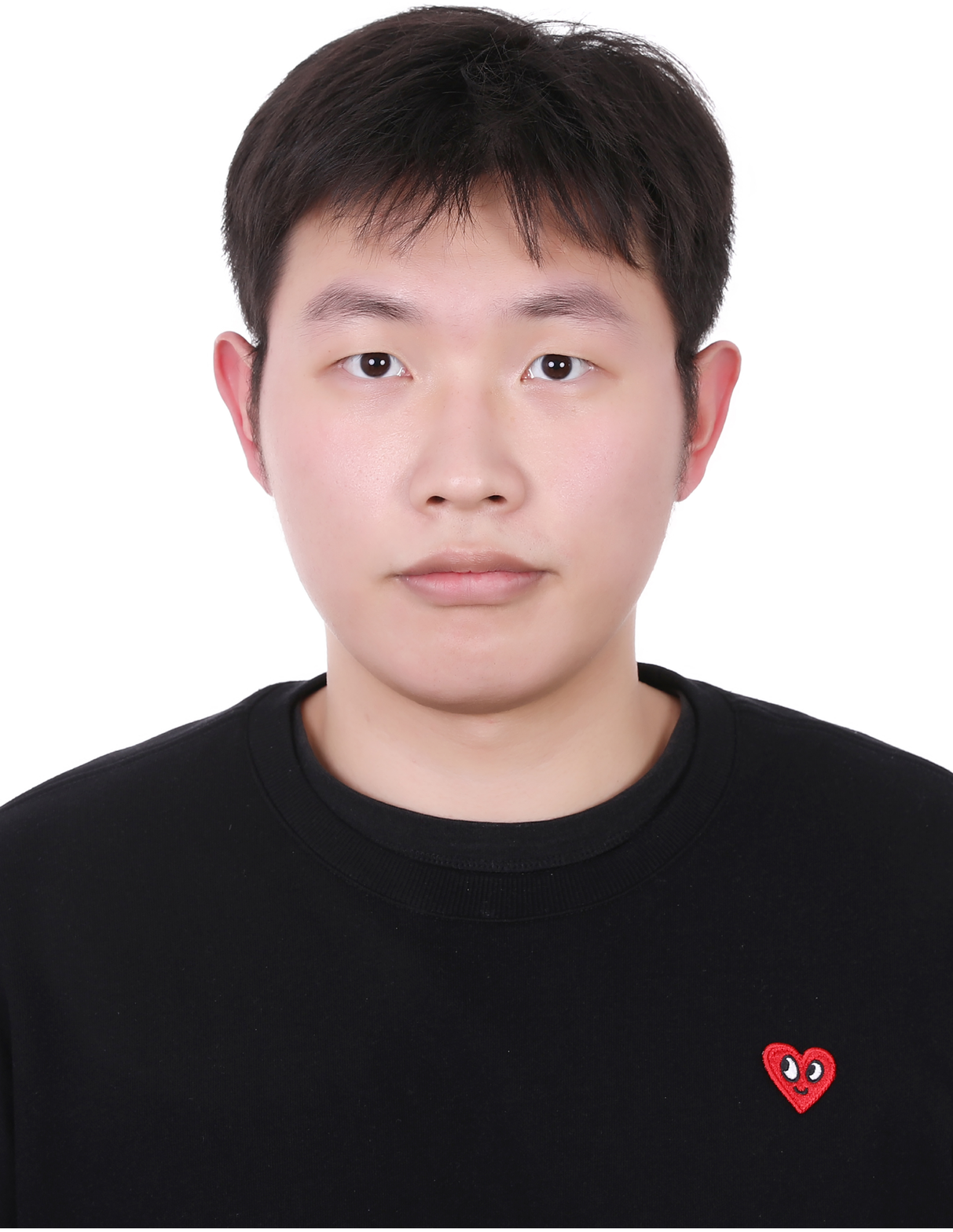}}]{Weiqi Sun} is currently a master candidate in School of Software, Beihang University. His current research interests include weakly-supervised temporal action localization and video retrieval. 
\end{IEEEbiography}
\vspace{11pt}

\begin{IEEEbiography}
[{\includegraphics[width=1in,height=1.25in,clip,keepaspectratio]{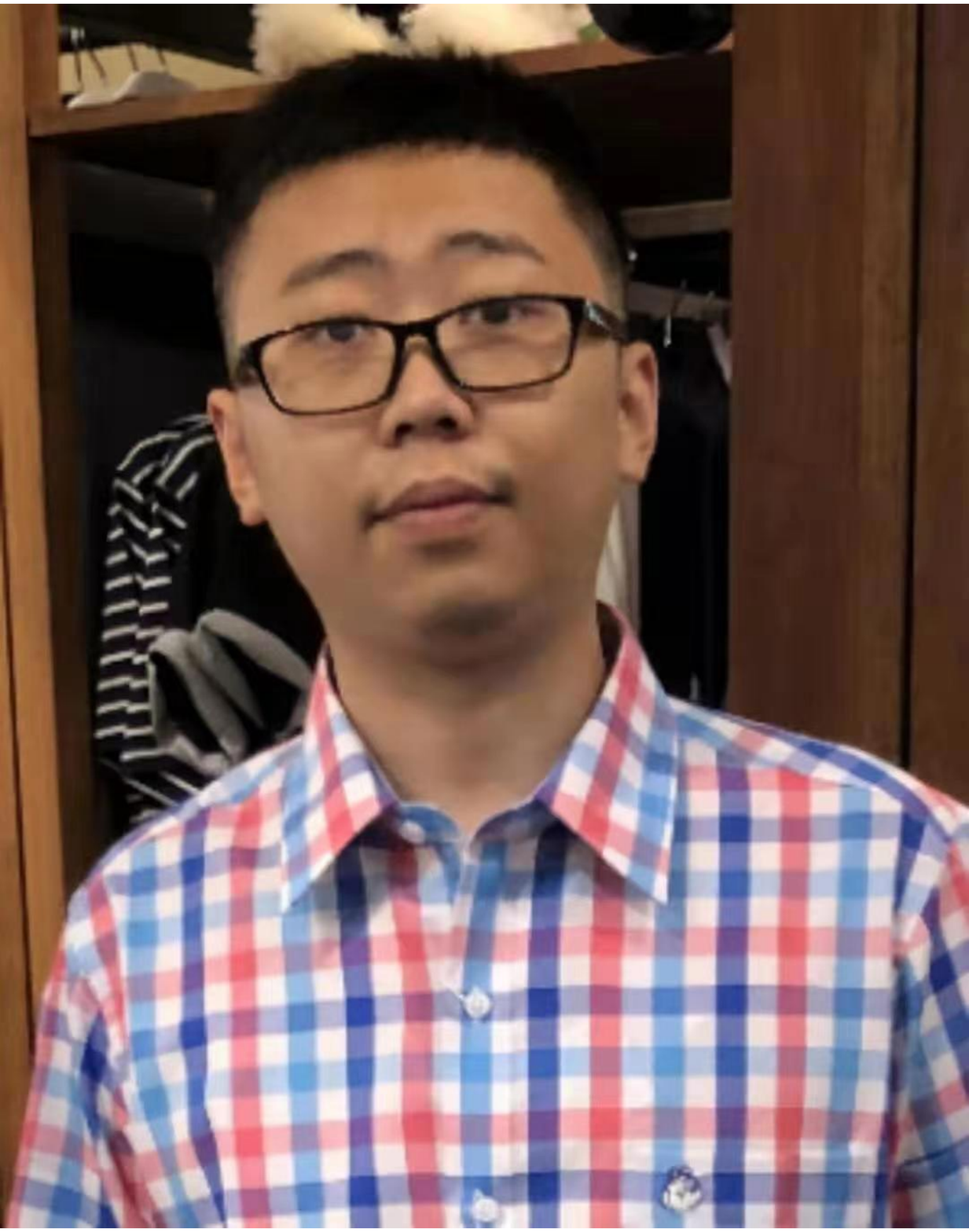}}]{Rui Su} received the B.S. degree in school of Electronic and Information Engineering from South China University of Technology in 2013,  the MPhil. degree in school of Information Technology and Electrical Engineering from the University of Queensland in 2016, and the PhD. degree in the school of Electrical and Information Engineering, the University of Sydney. His current research interests include action detection and its applications in computer vision.
\end{IEEEbiography}
\vspace{11pt}

\begin{IEEEbiography}
[{\includegraphics[width=1in,height=1.25in,clip,keepaspectratio]{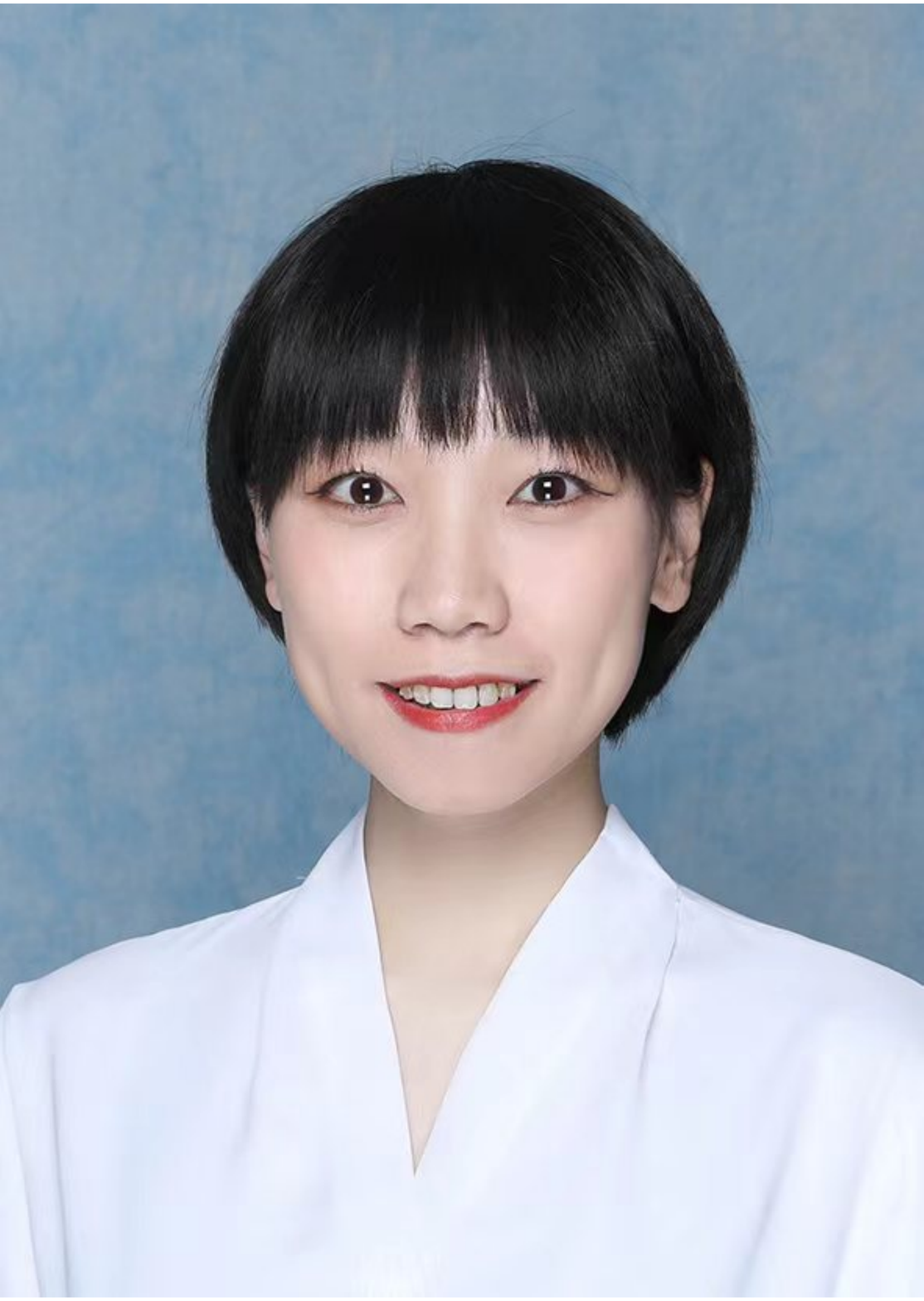}}]{Qian Yu} an associate professor at Beihang University. Before joining Beihang, she was a post-doctoral research scientist with UC Berkeley and International Computer Science Institute, working with Professor Stella Yu. She received her doctorate from the Queen Mary University of London, co-supervised by Professor Yi-Zhe Song and Professor Tao Xiang. Her research interest is sketch understanding and cross-modality modeling. Her research works have been published in top-tier journals and conferences including IJCV, CVPR, ECCV and ICCV. She was the recipient of Best Scientific Paper of BMVC 2015. She is one of the main organizers of ICCV 2021 workshop \textit{Sketching for Human Expressivity}.
\end{IEEEbiography}
\vspace{11pt}

\begin{IEEEbiography}
[{\includegraphics[width=1in,height=1.25in,clip,keepaspectratio]{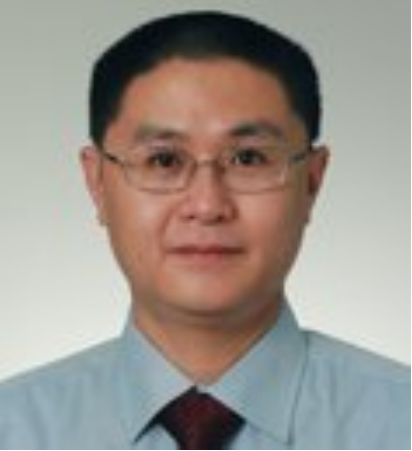}}]{Dong Xu} received the BE and PhD degrees from University of Science and Technology of China, in 2001 and 2005, respectively. While pursuing the PhD degree, he was an intern with Microsoft Research Asia, and a research assistant with the Chinese University of Hong Kong, for more than two years. He was a post-doctoral research scientist with Columbia University, for one year. He also worked as a Faculty Member at Nanyang Technological University, and the Chair of computer engineering at The University of Sydney. He is currently a Professor with the Department of Computer Science, The University of Hong Kong. His current research interests include computer vision, statistical learning, and multimedia content analysis. He was the co-author of a paper that won the Best Student Paper award in the IEEE Conference on Computer Vision and Pattern Recognition (CVPR) in 2010, and a paper that won the Prize Paper award in IEEE Transactions on Multimedia (T-MM) in 2014. He is a fellow of the IEEE.
\end{IEEEbiography}

\vspace{11pt}

\vfill

\end{document}